\documentclass[journal,10pt]{IEEEtran}

\usepackage{amsfonts}
\usepackage{amsmath}
\usepackage{amssymb}
\usepackage{graphicx}  %
\usepackage{multirow}
\usepackage{bm}
\usepackage{booktabs}
\usepackage{xspace}
\usepackage{xcolor}

\definecolor{blue}{rgb}{0,0,0}

\def\etal{{{\it et al.}}\xspace}
\def\eg{{\it e.g.}}

\usepackage[margin=7pt,font=footnotesize,labelfont=bf,labelsep=endash,tableposition=top]{caption}

\begin{document}

\title{Towards End-to-End Text Spotting in Natural Scenes}

\author{Peng Wang$^*$,
~~~
        Hui Li$^*$,
~~~
        Chunhua Shen
\thanks{
P. Wang is with  School of Computer Science, Northwestern Polytechnical University, China;
and National Engineering Laboratory for Integrated Aero-Space-Ground-Ocean Big Data Application Technology, China.
H. Li was with Samsung Research. C. Shen is with Monash University, Australia.
}
\thanks{
Work was done when
H. Li and C. Shen  were with
The University of Adelaide, Australia.
Correspondence should be addressed to C.
	Shen (e-mail: chunhua@me.com).
} %
\thanks{
$^*$ The first two authors equally contributed to this work.%
}
\thanks{
Accepted to
{\it IEEE Transactions on Pattern Analysis and Machine Intelligence (TPAMI),  June 2021.
Submitted June 2019.
}
}
}

\maketitle

\begin{abstract}

Text spotting in natural scene images is of great importance for many image understanding tasks. It includes two sub-tasks: text detection and %
recognition. In this work, we propose a unified network that simultaneously localizes and recognizes text with a single forward pass, avoiding intermediate processes such as  image cropping and feature re-calculation, word separation,  and
character grouping.  The overall framework is  trained end-to-end and is able to spot text of arbitrary shapes.  The convolutional features are calculated only once and shared by both the detection and recognition modules. Through multi-task training, the learned features become more {\color{black}{discriminative}} and improve the overall performance. By employing a $2$D attention model in word recognition, the issue of text irregularity is robustly addressed. The attention model
provides the spatial location for each character, which not only helps local feature extraction in word recognition, but also indicates an orientation angle to refine text localization. %
Experiments demonstrate
that our proposed method can achieve state-of-the-art performance on several widely-used text spotting benchmarks, including both regular and irregular datasets.

\end{abstract}


 \ifCLASSOPTIONpeerreview
 \begin{center} \bfseries EDICS Category: 3-BBND \end{center}
 \fi
\IEEEpeerreviewmaketitle

\section{Introduction}

\IEEEPARstart{T}{ext}---as a  fundamental  tool of communicating in\-for\-ma\-ti\-on---scatters throughout natural scenes, \eg, street signs, product labels, license plates, \textit{etc}.
Automatically reading text in natural scene images is an important task in machine learning %
and gains increasing attention due to a variety of
applications. For example, accessing text in images can help
the visually impaired
understand the surrounding  environment.
To enable autonomous driving, one must accurately detect and recognize every road sign.
Indexing text in images would enable image search and retrieval from billions of consumer photos in internet.

End-to-end text spotting includes two sub-tasks: text detection and word recognition. Text detection aims to
localize
each
text in images, %
using
bounding boxes for example.
Word recognition attempts to output %
readable  transcription.
Compared to traditional optical character recognition (OCR), text spotting in natural scene images is
much
more challenging,
mainly due to
the extreme diversity of text patterns and highly complicated background. Text appearing in natural scene images can be of varying fonts, sizes, shapes, orientation and layouts. %
Moreover, the background can be cluttered, making the task largely unsolved to date.

An intuitive approach to  scene text spotting is to divide it into two separated sub-tasks. Text detection is first
performed
to obtain  candidate text bounding boxes, and word recognition is %
applied
subsequently on the cropped regions to  output   transcriptions. %
A few
approaches %
were
developed which solely focus on text detection~\cite{Yin2014pami, EAST17, liu17, Liao2018Text, Dingerrui19} or word recognition~\cite{ShiBY17, Cheng2018AON, shiPAMI2018, ESIR19}. Methods are improved from only %
recognizing
simple horizontal text to addressing complicated irregular (oriented or curved) text. %
We believe that
these two sub-tasks are highly correlated and complementary
to each other, and thus should be solved in a single framework.
On one hand, %
image
features
can
be shared between these two tasks so as to %
reducing computational cost.
On the other hand, the multi-task training %
is likely to
improve feature representation power and benefit both sub-tasks.

\begin{figure*}[t!]
	\begin{center}
		\includegraphics[width=0.96\textwidth]{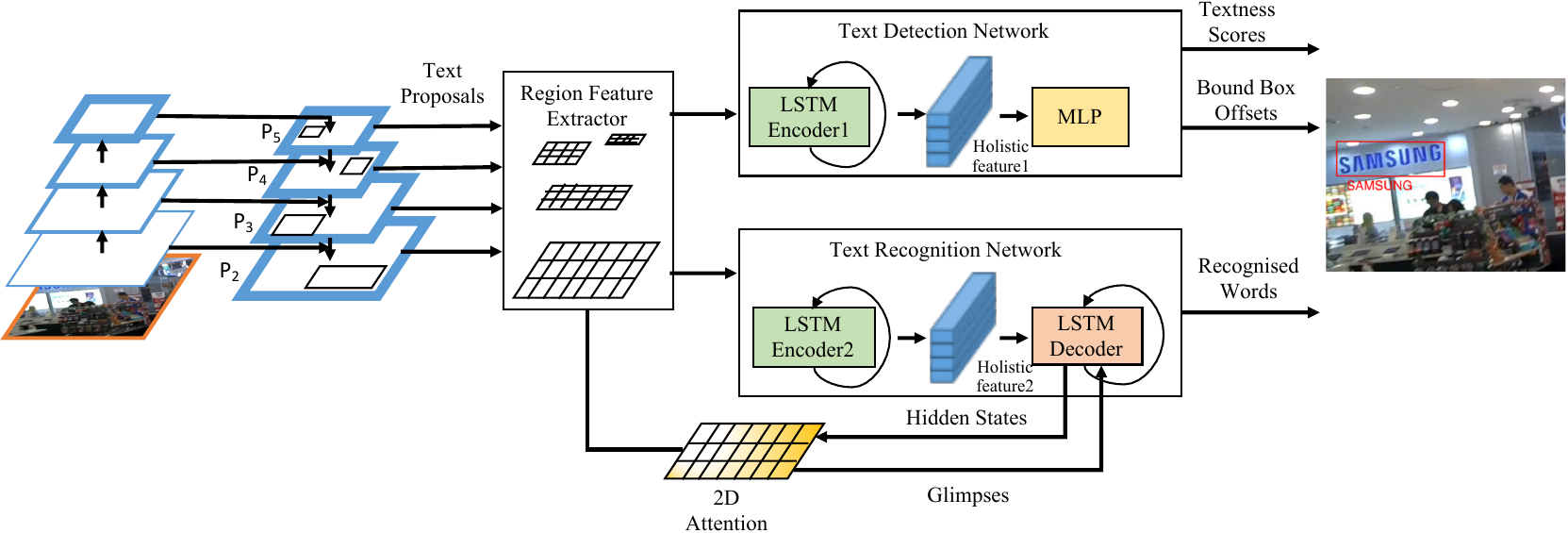}
	\end{center}
	\caption{\textbf{The overall architecture of our proposed model for end-to-end text spotting in natural scene images}. The network takes an image as input, and outputs both text bounding boxes and text labels in one single forward pass. The entire network is trained end-to-end.}
	\label{fig:overall}
\end{figure*}

To this end, %
end-to-end approaches are proposed recently to concurrently tackle both sub-tasks~\cite{li2017towards,hetong2018,FOTS2018,Maskspot2018}.
Note
that most end-to-end approaches %
spend major effort
on designing  %
sophisticated detection modules, so as to acquire tighter bounding boxes around the text, %
alleviating the %
level of difficulty
for word recognition.
However, we argue that
the ultimate goal of text spotting is to
recognize each text in the image,
rather than
attaining
precise
bounding box locations.
Thus, here we strive for a balance between detection and recognition by
letting the recognition module deal with the challenge caused by text irregularity.
To be more specific,
our
detection module is designed to output a rectangular bounding box for each word,
regardless of
what
the text appears
(horizontal, oriented or curved).  A robust recognition module, which shares image features with the detection module, is devised to
effectively
recognize the text within the relatively loose bounding box.  The overall framework of our method is %
illustrated
in Figure~\ref{fig:overall}.
We use the
ResNet~\cite{resnet} as the backbone, with Feature Pyramid Networks (FPN)~\cite{FPN} %
used to tackle the multi-scale detection.
Text Proposal network (TPN) is adapted
at
multiple levels
of
the feature pyramid so as to obtain text proposals %
of
different scales. A RoI pooling layer is then employed to extract varying-size $2$D features from each proposal, which are then %
used both in the text detection network and word recognition network. A $2$-dimensional attention network is %
employed
in the word recognition module.
The 2D attention model attends on
the local discriminative features for each individual character during decoding,
which
improves the recognition accuracy.
At the same time, the attention model
indicates the character alignment in each word bounding box, with which
we can
refine the loosely localized bounding box. The recognition module can also %
be used to
reject false positives in the detection phase, thus
improving the overall performance.

This work here is an extension of our previous %
work published in~\cite{li2017towards} and~\cite{li2019aaai}. The work in Li~\etal \cite{li2017towards} proposed {\it the first end-to-end trainable framework for scene text spotting}.
However,
a significant drawback of \cite{li2017towards} is that
it is incapable of dealing with
irregular text that is oriented or curved. In \cite{li2019aaai}
we presented
a $2$D attention based simple baseline for irregular text recognition, which is robust to various distortions. Here we
still
use the 2D attention based encoder-decoder framework in the recognition branch. Moreover, the calculated attention weights are
exploited
to compute the orientation angle for bounding box refinement in the end-to-end scenario.
The improvements compared to~\cite{li2017towards} are as follows.
\begin{enumerate}

\item

 The work here is able to tackle text with arbitrary shapes. It is no longer restricted %
 to
 horizontal text as in~\cite{li2017towards}.

 \item

 We now use  ResNet with FPN as the backbone network,
 leading to significantly
 improved
 feature representations. We also adapt the text proposal network with pyramid feature maps. These two modifications are able to propose text instances at a wide range of scales and improve the recall %
 for
 small size text.

 \item The training process is %
 made much simpler.
 Instead of training the detection and recognition modules separately at the early stages as in~\cite{li2017towards}, the new framework is trained completely in a %
 simple
 end-to-end fashion. Both detection and recognition tasks are jointly optimized in the %
 training process.
 Our code is optimized, resulting in a faster computational speed compared to~\cite{li2017towards}.

 \item  More experiments are conducted on three additional datasets to demonstrate the effectiveness of the proposed method in dealing with  various text appearance.
\end{enumerate}

The main contributions of this work are thus  four-fold.
\begin{enumerate}
    \item

 We design an end-to-end trainable network, %
 which can localize text in natural scene images and recognize it simultaneously. %
 The method is robust to %
 text appearance variation %
 in that
 it can %
 detect and recognize
 arbitrarily-oriented text.
 The convolutional features are shared by both
 the
 detection and recognition modules, which %
 reduces
 computational cost,
 comparing with approaches that need
 two distinct models. In addition, the multi-task optimization  benefits the feature learning, and thus promotes the detection results
 and
 the overall performance. To our  knowledge, ours is the first work that integrates text detection and recognition into a single end-to-end trainable network.

\item A tailored RoI pooling method is proposed, which takes %
the significant diversity of aspect ratios in text bounding boxes
into account. The generated RoI feature maps accommodate the aspect ratios of different words and %
extract
sufficient information
that is essential for the subsequent detection and recognition.

\item We take full use of the $2$D attention mechanism in both word recognition and bounding box refinement. The learned attention weights can not only select local features to boost recognition performance, but also provide character locations %
which can be used to
refine the bounding boxes.
Note that
the $2$D attention model is trained in a
\textit{weakly supervised} manner using the cross-entropy loss in word recognition. We do not require additional pixel-level or character-level annotations for supervision.

\item Our work provides a new %
approach
to solving  the end-to-end text spotting problem.
Conventional methods have been built on the idea of  %
obtaining
accurate and tight bounding boxes around the text
at the first step, so as to exclude redundant noises and %
make the
word recognition task easier.
In contrast, a strong and robust word recognition model underpins our framework,
which can compensate the detection module, leading to a simple end-to-end text spotting method.

Experiments on several widely-used text spotting benchmarking datasets,
including ICDAR2013, ICDAR2015, Total-Text and COCO-Text, achieve
state-of-the-art results, demonstrating the effectiveness of our method.

\end{enumerate}

\section{Related Work}
In this section, we %
review
some related work on text detection, word recognition and end-to-end text spotting. %
Comprehensive surveys %
on
scene text detection and recognition can be found in  \cite{survey18, Ye2015pami, xiang2016Survey, liu2016tip}.

{\bf Text Detection}
With the development of deep learning,
text detection in natural scene images has achieved  significant progress.
Methods have been developed for
detecting regular horizontal text, %
oriented
and
curved text.

Pioneering methods
such as
\cite{Max2014ECCV, Weilin2014ECCV} simply use pre-trained Convolutional Neural Networks (CNNs) as classifiers to distinguish characters from background. Heuristic steps are needed to group characters into words. Zhang~\etal~\cite{zhengCVPR15} proposed to extract text lines by exploiting text symmetry property.
Tian~\etal~\cite{Tian2016} developed a vertical anchor mechanism, and proposed a Connectionist Text Proposal Network (CTPN) to accurately localize text lines in images. %
Advances in generic
object detection and segmentation provide %
inspirations for text detection.
For example, inspired by Faster-RCNN~\cite{renNIPS15fasterrcnn}, Zhong~\etal~\cite{Zhong2016} designed a text detector with a multi-scale Region Proposal Network (RPN) and a multi-level RoI pooling layer which can localize word level bounding boxes directly. Gupta~\etal~\cite{Gupta16} used a Fully-Convolutional Regression Network (FCRN) for efficient text detection and bounding box regression, motivated by YOLO~\cite{YOLO2016}. Similar to SSD~\cite{SSD2016}, Liao~\etal~\cite{LiaoAAAi2017} proposed ``TextBoxes'' by combining predictions from multiple feature maps with different resolutions. Those methods
output horizontal rectangles for detecting  text of regular shapes.

The authors of \cite{Zhang_2016_CVPR}
proposed to localize text lines via salient maps.
Post-processing techniques  were  proposed to extract text lines in multiple orientations. Ma~\etal~\cite{RRPN} introduced Rotation Region Proposal Networks (RRPN) to generate
orientated
proposals.
He~\etal~\cite{He2017ICCV} proposed to use an attention mechanism to identify text regions from images.
The bounding box position was regressed with an angle for box orientation. These methods  output rotated rectangular bounding boxes. In addition, Zhou~\etal~\cite{EAST17} proposed ``EAST" that %
{\color{black}{uses}} FCN to produce word %
level predictions which can be either rotated rectangles or quadrangles. Liu~\etal~\cite{liu17} proposed Deep Matching Prior Network (DMPNet) to detect text with tighter quadrangle.
Liao~\etal~\cite{Liao2018Text} improved ``TextBoxes'' to
predict
orientation angles or quadrilateral bounding box offsets so as to detect oriented scene text (referred  to as ``TextBoxes++'').
 Lyu~\etal~\cite{Lyucvpr18} proposed to detect scene text by localizing the corner points of text bounding boxes and segmenting text regions in relative positions. Candidate boxes are generated by sampling and grouping corner points, which results in quadrangle detection.

 Recently, more advanced methods are proposed to predict bounding boxes of polygons which aim to fit text
 more tightly.
 For example, inspired by Mask R-CNN~\cite{maskrcnn}, Xie~\etal~\cite{xieAAAI19} proposed to detect arbitrary shape text based on FPN~\cite{FPN} and instance segmentation.
 Zhang~\etal~\cite{Dingerrui19} proposed to detect text via iterative refinement and shape expression. An instance-level shape expression module was introduced to generate polygons that can fit arbitrary-shape text (\eg, curved).  Progressive Scale Expansion Network (PSENet)~\cite{PSENet19}
 performs pixel-level segmentation for
 localizing
 text instances {\color{black}{precisely}} of  arbitrary shapes.
 Tian~\etal~\cite{jiaya19}
 solved
 text detection %
 using
 instance segmentation. Pixels belonging to the same word are pulled together as connected components while pixels from different words are pushed away from each other. %

 The text detection  module of our framework  is %
 based on the Faster R-CNN framework~\cite{renNIPS15fasterrcnn},  which aims to generate word-level bounding boxes directly,  eliminating intermediate steps such as character aggregation and text line separation.  In order to cover text %
 of
 a variety of scales and aspect ratios, FPN~\cite{FPN} is %
 used
 to generate text proposals with both higher recall and precision. {\color{black}{Different from other works, we %
 use
 the minimal horizontal rectangle that encloses the whole word as the ground-truth.}} %
It contains sufficient information %
for
text spotting. Besides, the %
overall
framework can be simplified as we do not need additional modules to %
accommodate
text orientation. In our framework, a %
preciser localization can be obtained %
using the
word recognition results.

{\bf Word Recognition}
Word recognition means to recognize the cropped word image patches %
and outputs
character sequences.
Early work %
on
scene text recognition often works
in a
bottom-up fashion~\cite{Wangkai2011,Max2014ECCV}, which detects individual characters firstly and integrates them into a word by %
dynamic programming.
 Top-down %
 methods
 \cite{maxNIPS14}, %
 recognize a
 word patch as a whole %
 formulate as a
 multi-class %
 classification problem. Considering that scene text generally appears in the form of a character sequence, recent work models it as a sequence recognition problem. Recurrent Neural Networks (RNNs) are %
 employed for this purpose.

The work in~\cite{He2015Reading} and \cite{ShiBY17} %
formulates
word recognition as one-dimensional sequence labeling problem using RNNs.
A Connectionist Temporal Classification (CTC) layer~\cite{Graves2006ICML} is %
used
to decode the %
sequences, eliminating the need of segmenting characters. %

The works
in~\cite{Lee_2016_CVPR} and \cite{shiCVPR2016} %
recognize text using an attention-based sequence-to-sequence framework~\cite{NIPS2014},
in which
RNNs are able to learn the character-level language model.
A $1$D soft-attention model was employed to select relevant local features.
The RNN+CTC and sequence-to-sequence frameworks serve as two meta-algorithms that are widely %
used
by
recent
approaches.
Both models can be trained end-to-end and achieve considerable improvements on regular text recognition.
Cheng~\etal~\cite{cheng_EditDistance} observed that the frame-wise maximal likelihood loss, which is conventionally used to train the encoder-decoder framework, may be
confused and misled by missing or superfluity of characters, and {\color{black}{degrades}} the recognition accuracy.
They proposed ``Edit Probability'' to %
tackle
this misalignment problem.

Recognizing irregular text has also attracted much attention recently.
Shi~\etal~\cite{shiPAMI2018, shiCVPR2016} rectified oriented and curved text
using
Spatial Transformer Network (STN)~\cite{jaderberg2015spatial}.
ESIR~\cite{ESIR19} employed a line-fitting transformation to estimate the pose of text, and developed a pipline that iteratively removes perspective distortion and text line curvature%
in order to achieve improved recognition accuracy.

Instead of rectifying the whole distorted text image, Liu~\etal~\cite{Liu2018CharNetAC} presented a Character-Aware Neural Network (Char-Net) to detect and rectify individual characters, which, however, requires %
expensive
character-level annotations.
Cheng~\etal~\cite{Cheng2017} proposed a Focusing Attention Network (FAN) that is composed of an attention network for character recognition and a focusing network to adjust the attention drift between local character features and targets. Character-level bounding box annotations {\color{black}{are}} also %
required.
Cheng~\etal~\cite{Cheng2018AON} applied LSTMs in four directions to encode arbitrarily-oriented text. %
The work in~\cite{li2019aaai} depends on a tailored $2$D attention mechanism to deal with the complicated spatial layout of irregular text, and shows significant flexibility and robustness. In this work, we %
use a 2D attention model
in the recognition module, and train altogether with the detection
module for  end-to-end text spotting.

{\bf End-to-End Text Spotting}
Most previous methods design a multi-stage pipeline to achieve text spotting. For instance, Jaderberg~\etal~\cite{Max2016IJCV} generated a %
large number
of text proposals,%
and then
trained
the word classifier
for recognition.  Gupta~\etal~\cite{Gupta16} employed FCRN for text detection and the word classifier in~\cite{maxNIPS14} for recognition. Liao~\etal~\cite{Liao2018Text} combined ``TextBoxes++'' and ``CRNN''~\cite{ShiBY17} to complete the text spotting task. The work in ASTER~\cite{shiPAMI2018} combines ``TextBoxes''~\cite{LiaoAAAi2017} and a rectification based recognition method for text spotting.

Preliminary results of this work, presented in \cite{li2017towards}, %
is
among the first a few, %
to explore a unified end-to-end trainable framework for %
simultaneous
text detection and recognition. Although in one single framework, the work in~\cite{Textspotter_ICCV2017} does not share %
features between detection and recognition parts, which
can be seen as
a loose combination. Our previous work~\cite{li2017towards} shares the RoI features for both detection and recognition for reducing computation, %
{\color{black}{Meanwhile}}, the joint optimization of multi-task loss can also improve
the overall performance.
He~\etal~\cite{hetong2018} proposed an end-to-end text spotter which can compute convolutional features for oriented text instances. A $1$D character attention mechanism was introduced via explicit alignment which improves performance. Note that
character level annotations are needed for supervision. %
Liu~\etal~\cite{FOTS2018} presented ``FOTS'' that proposes ``RoIRotate'' to share convolutional features between detection and recognition for oriented text. $1$D sequential features are extracted via several layers of CNNs and RNNs, and decoded by a CTC layer. Both work may encounter difficulty in dealing with curved or distorted scene text, %
where the orientation is well defined.

Lyu~\etal~\cite{Maskspot2018} proposed ``Mask TextSpotter'' that introduced a mask branch for character instance segmentation, inspired by Mask R-CNN~\cite{maskrcnn}.
It can detect and recognize text of various shapes, including horizontal, oriented and curved text. %
Again,
character-level mask information is needed for training. {\color{black}{Its extended version~\cite{LiaoPAMI2019} integrated a Spatial Attention Module in the recognition %
module,
which mitigates the above-mentioned problem.}}
Sun~\etal~\cite{Sun2018TextNetIT} proposed ``TextNet'' to read irregular text. It outputs quadrangle text proposals. A perspective RoI transform is developed to extract features from arbitrary-size quadrangle for recognition. Four directional RNNs are %
used
to encode the irregular text instances, {\color{black}{which results in}} context features for the following spatial attention mechanism in the decoding process. {\color{black}{More recently, Qin~\etal~\cite{Qin2019ICCV} formulated arbitrary shape text spotting as an instance segmentation problem.
Xing~\etal~\cite{Xing2019ICCV} proposed convolutional character networks,
which detects and recognizes at the character level.
Liu~\etal~\cite{ABCNet} proposed ABCNet that %
fits
cubic Bezier curves to curved text and designs a BezierAlign layer to extract curved sequence features.}}

In contrast to designing a sophisticated framework to
accommodate
the variety of text shapes, which can potentially increase  model complexity, we resort to the conventional horizontal bounding box {\color{black}{to represent}} text location.
It not only provides sufficient information to %
fulfill
the
text spotting task, but also leads to a considerably simpler model.
We %
postpone
the processing of text irregularity  to the flexible yet strong $2$D attention model in word recognition. %
{
\color{blue}{%
The work of
ASTER~\cite{shiPAMI2018} used the control points obtained in recognition model to rectify the detection results, which is somehow similar to our post-processing step.
    The main difference is that our model is end-to-end trainable and ASTER is not.
  }}

\section{Our Method}

The overall architecture of our proposed model is illustrated in Figure~\ref{fig:overall}. Our goal is to design an end-to-end trainable network, which can simultaneously detect and recognize all words in natural scene images, %
robust to
various appearances. The overall framework consists of five  components:
1) a ResNet backbone %
with FPN embedded for feature extraction;
2) a TPN with a shared head across all feature pyramid levels for text proposal generation;
3) a Region Feature Extractor (RFE) to extract varying length $2$D features that accommodate text aspect ratios and are shared by following detection and recognition modules;
4) a Text Detection Network (TDN) for proposal classification and bounding box regression; and
5) meanwhile a Text Recognition Network (TRN) with $2$D attention equipped for proposal recognition.

We attempt to design a simple model.
Hence,  we exclude additional modules for %
dealing with
the irregularity of text. Instead, we solely
rely on a $2$D attention mechanism in both word recognition and location refinement.  Despite its simplicity, we shown that our mode is robust
in various scenarios.
In the following, we describe each %
component
of the model in detail.

\subsection{Backbone}

A pre-trained ResNet~\cite{resnet}  is used
here as the backbone convolutional layers for its
good
performance on image recognition. It consists of $5$ residual blocks with down sampling ratios of $\{ 2,4,8,16,32 \}$ separately for the last layer of each block, with respect to the input image. We remove the final pooling and fully connected layer. Thus an input image %
leads to
a pyramid of feature maps. In order to build high-level semantic features, FPN~\cite{FPN} is applied which uses a bottom-up and a top-down pathways with lateral connections to learn a strong semantic feature pyramid at all scales. It shows a significant improvement on bounding box proposals~\cite{FPN}. Similarly, we exclude the output from conv1 in the feature pyramid, and denote the final set of feature pyramid maps as $\{ P_2, P_3, P_4, P_5 \}$. The feature dimension is also fixed to $d=256$ in all feature maps.

\subsection{Text Proposal Network}
In order to take full use of the rich semantic feature pyramid as well as the location information, following the work in~\cite{FPN}, we attach a head with $3 \times 3$ convolution and two sibling $1 \times 1$ convolutions (for text/non-text classification and bounding box regression respectively) to each level of the feature pyramid, which gives rise to anchors at different levels. Considering the relatively small sizes of text instances, we define the anchors of sizes $ \{ 16^2, 32^2, 64^2, 128^2, 256^2 \}$ pixels on $\{ P_2, P_3, P_4, P_5 , P_6\}$ respectively, where $P_6$ is a stride two subsampling of $P_5$. The aspect ratios are set to $\{ 0.125, 0.25, 0.5, 1.0 \}$ by considering that text bounding boxes usually have larger width than height. Therefore, there are
in total
$20$ anchors over the feature pyramid, which are capable of covering text instances with different {\color{black}{scales and}} shapes.

The heads with $3 \times 3$ conv and two $1 \times 1$ conv's share parameters across all feature pyramid levels. They extract features with $256$-d from each anchor and fed them into two sibling layers for text/non-text classification and bounding box regression. The training of TPN follows the work in FPN~\cite{FPN}.

\subsection{Region Feature Extractor}
Given that text instances usually have a large variation on word length, it is unreasonable to make fixed-size RoI pooling for short words like ``Dr'' and long words like ``congratulations''. This would inevitably  lead to significant distortion in the produced feature maps,
which is disadvantageous  for the downstream  text detection and recognition networks. In this work, we propose to re-sample regions according to their perspective aspect ratios. RoI-Align~\cite{maskrcnn} is also used to improve alignment between input and output features. For RoIs of different scales, we assign them to different pyramid levels for feature extraction, following the method in~\cite{FPN}. The difference is that, for an RoI of size $h \times w$,  a spatial RoI-Align is performed with the resulting feature size of
\begin{equation}
H \times \max ( H, \min(W_{max},3Hw/h) ),
\label{eq:pool}
\end{equation}
where the expected height $H$ is fixed to $4$, and the width is adjusted to accommodate the large variation of text aspect ratios. The resulted feature maps are  denser along the width direction compared to the height direction, which reserves more information along the horizontal axis and benefits the
subsequent
recognition task. Moreover, the feature width is clamped by $H$ and a maximum length $W_{max}$ which is set to $30$ in our work. The resulting  $2$D feature maps (denoted as $\mathbf{V}$ of size $H \times W \times D$ where $D=256$ is the number of channels) are used: 1) to extract holistic features for the following text detection and recognition; 2) as the context for the $2$D attention network in text recognition.

\subsection{Text Detection Network}
Text Detection Network (TDN) aims to %
classify
whether the proposed RoIs are text or not and refine the coordinates of bounding boxes %
again, based on the extracted region features $\mathbf{V}$. Note that $\mathbf{V}$ is of varying sizes. To extract a fixed-size holistic feature from each proposal, RNNs with Long-Short Term Memory (LSTM) is adopted. We flatten the features in each column of $\mathbf{V}$, and obtain a sequence $\{ \mathbf{q}_1, \dots, \mathbf{q}_W \}$ where $ \mathbf{q}_t \in \mathbb{R}^{D \times H}$.  The sequential elements are fed into LSTMs one by one. Each time LSTMs receive one column of feature $\mathbf{q}_t$, and update their hidden state $\mathbf{h_d}_t$ by a non-linear function: $\mathbf{h_d}_t =\mathrm{f} (\mathbf{q}_t, \mathbf{h_d}_{t-1})$. In this recurrent fashion, the final hidden state $\mathbf{h_d}_W$ (with size $R = 1024$) captures the holistic information of $\mathbf{V}$ and is used as a RoI representation with fixed dimension. Two fully-connected layers with $1024$ neurons are applied on $\mathbf{h_d}_W$, followed by two parallel layers for classification and bounding box regression respectively.

To boost the detection performance, an online hard negative mining is
used
during  training.
We firstly apply TDN on $1024$ initially proposed RoIs. The ones that have higher textness scores but are actually negatives are re-sampled to harness TDN. In the re-sampled RoIs, we restrict the positive-to-negative ratio as $1:3$, where in the negative RoIs, we use $70 \%$ hard negatives and $30 \%$ random sampled ones. Through this %
processing, we observe that
the text detection performance can be improved
significantly.

\subsection{Text Recognition Network}
Text Recognition Network (TRN) aims to predict the text in the detected bounding boxes based on the extracted region features. Considering the irregularity of text, we apply a $2$D attention mechanism based the encoder-decoder network for text recognition, %
following
the work in~\cite{li2019aaai}. The extracted RoI feature $\mathbf{V}$ is adopted directly in the recognition network, instead of cropping the text proposals out and feeding to another standalone backbone CNNs for feature extraction.
 Without additional transformation on the extracted RoI features, the proposed attention module is able to accommodate text of arbitrary shape, layout and orientation.

The extracted RoI feature $\mathbf{V}$ is encoded again to extract {\color{black}{discriminative}} features for word recognition.
$2$ layers of LSTMs are employed here in the encoder, with $512$ hidden states per layer.
The LSTM encoder receives one column of the $2$D features maps at each time step,  followed by max-pooling along the vertical axis, and updates its hidden state $\mathbf{h}_t$.
After $W$ steps, the final hidden state of the second RNN layer, $\mathbf{h}_{W}$, is regarded as the holistic feature for word recognition.

\begin{figure}[t]
	\begin{center}
		\includegraphics[width=0.45\textwidth]{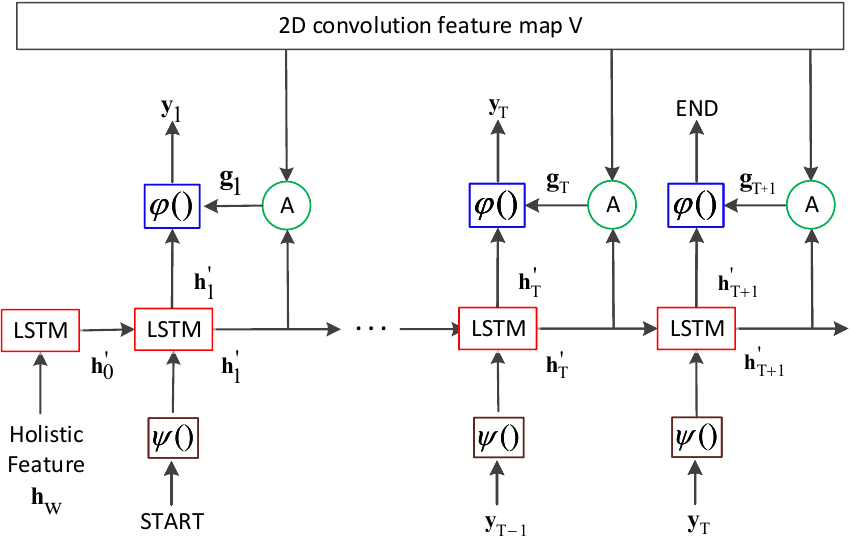}
	\end{center}
	\caption{ The structure of the LSTM decoder used in this work. The holistic feature $\mathbf{h}_{W}$, a ``START'' token and the previous outputs are input into LSTM subsequently,
		terminated by an ``END'' token. At each time step $t$, the output $y_t$ is computed by $\varphi(\cdot)$ with the current hidden state and the attention output as inputs.
	}
	\label{fig:decoder}
\end{figure}

The decoder is another $2$-layer LSTMs with $512$ hidden states per layer.
Here the encoder and decoder do not share parameters.
As illustrated in Figure~\ref{fig:decoder}, initially, the holistic feature $\mathbf{h}_{W}$ is fed into the decoder LSTMs at time step $0$.
Then a ``$\rm START$'' token is input into LSTMs at step $1$.
From time step $2$,
the output of the previous step is fed into LSTMs until the ``$\rm END$'' token is received.
All the inputs to LSTMs are represented by one-hot vectors, followed by a linear transformation $\Psi(\cdot)$.

During training, the inputs of decoder LSTMs are replaced by the ground-truth character sequence.
The outputs are computed by the following transformation:
\begin{equation}
\mathbf{y}_t = \mathrm{ \varphi( \mathbf{h}'_t, \mathbf{g}_t   )} = \mathrm{softmax}(\mathbf{W}_o  [ \mathbf{h}'_t; \mathbf{g}_t ] )
\end{equation}
where $\mathbf{h}'_t$ is the current hidden state and $\mathbf{g}_t$ is the output of the attention module.
$\mathbf{W}_o $ is a linear transformation, which embeds features into the output space of $38$ classes, in corresponding to $10$ digits, $26$ case insensitive letters, %
one special token representing all punctuation, and an ``END'' token.

The attention model $\mathbf{g}_t = \mathrm{Atten}(\mathbf{V},\mathbf{h}'_{t})$ is defined as:
\begin{equation}
\begin{cases}
\mathbf{e}_{ij} = \tanh( \mathbf{W}_v \mathbf{v}_{ij} + \mathbf{W}_h \mathbf{h}'_t), \,\, \\

{\alpha}_{ij} = \mathrm{softmax} (\mathbf{w}_e^T \cdot \mathbf{e}_{ij}), \\
\mathbf{g}_t = \!\displaystyle{\sum_{i,j}} \, \alpha_{ij} \mathbf{v}_{ij},  \quad  i = 1, \dots, H, \quad j = 1, \dots, W.
\end{cases}
\label{eq:atten}
\end{equation}
where $\mathbf{v}_{ij}$ is the local feature vector at position $(i,j)$ in the extracted region feature $\mathbf{V}$;
$\mathbf{h}'_t$ is the hidden state of decoder LSTMs at time step $t$, to be used as the guidance signal;
$\mathbf{W}_v$ and $\mathbf{W}_h$
are linear transformations to be learned; ${\alpha}_{ij}$ is the attention weight at location $(i,j)$; and $\mathbf{g}_t$ is the weighted sum of local features, denoted as a \textit{glimpse}.

\begin{figure}[!th]
	\begin{center}
		\includegraphics[width=0.45\textwidth]{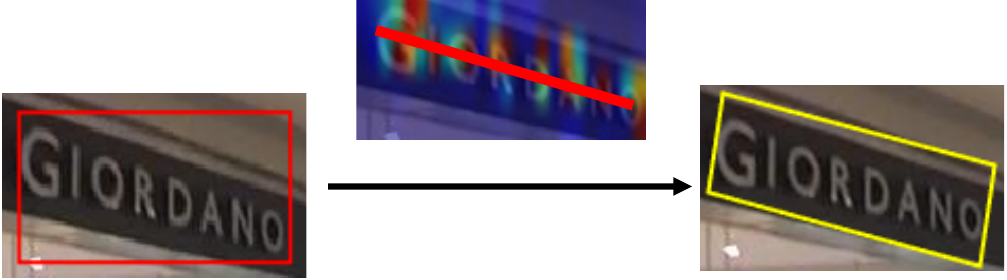}
	\end{center}
	\vspace{-3mm}
	\caption{Box refinement according to character alignment indexed by attention weights.
	}
	\label{fig:boxrotate}
\end{figure}

The attention module is learned in a \textit{weakly supervised} manner by the cross entropy loss in the final word recognition. \textit{No pixel-level or character-level annotations are required for supervision in our model}.
The calculated attention weights can not only extract {\color{black}{discriminative}} local features for the character being decoded and help word recognition, but also provide a group of character location information. For irregular text, an orientation angle is then calculated based on the character locations in the proposal, which can be used to refine the bounding boxes afterwards. To be more specific, as shown in Figure~\ref{fig:boxrotate}, a linear equation can be regressed based on the character locations specified by the attention weights in decoding process. The output rectangle is then rotated based on the computed  slope. In practice, we remove attention weights smaller than $0.2$ to reduce noise.

\subsection{Loss Functions and Training}

Our proposed framework is trained in an end-to-end manner, requiring only input images, the ground-truth word bounding boxes and their text labels as input during training phase. Instead of requiring quadrangle or more sophisticated polygonal coordinate annotations, in this work we %
are able to
use the simplest horizontal bounding box which indicates the minimum rectangle encircling the word instance. In addition, no pixel-level or character-level annotations are requested for supervision. Specifically, both TPN and TDN employ the binary logistic loss $L_{{cls}}$ for classification, and smooth $L_1$ loss $L_{{reg}}$~\cite{renNIPS15fasterrcnn} for regression.
So the loss for training TPN is
\begin{equation}\label{eq1}
L_{{T\!P\!N}}=\frac{1}{N} \sum_{i=1}^{N} {L}_{{cls}} (p_i,p_i^\star)  + \frac{1}{N_{+}}
\sum_{i=1}^{N_+} L_{{reg}} (\mathbf{d}_i,\mathbf{d}_i^\star),
\end{equation}
where $N$ is the number of randomly sampled anchors in a mini-batch and $N_+$ is the number of positive anchors in this batch.
The mini-batch sampling and training process of TPN are similar to that used in~\cite{FPN}.

An anchor is considered as positive if its Intersection-over-Union (IoU) ratio with a ground-truth
is greater than $0.7$ and considered as negative if its IoU with any ground-truth is
smaller than $0.3$.
$N$ is set to $256$ and $N_+$ is at most $128$.
$p_i$ denotes the predicted probability of anchor $i$ being text and $p_i^\star$
is the corresponding ground-truth label ($1$ for text, $0$ for non-text).
$\mathbf{d}_i$ is the predicted coordinate offsets $(\mathrm{dx}_i, \mathrm{dy}_i, \mathrm{dw}_i, \mathrm{dh}_i)$ for anchor $i$, which indicates scale-invariant translations and log-space height/width shifts relative to the pre-defined anchors,
and $\mathbf{d}_i^\star$ is the associated offsets for anchor $i$ relative to the ground-truth.
Bounding box regression is only for positive anchors, as there is no ground-truth bounding box matched with negative ones.

For the final outputs of the whole system, we apply a multi-task loss for both detection and recognition:
\begin{align}\label{eq2}
L_{D\!R\!N} &= \frac{1}{\hat{N}} \sum_{i=1}^{\hat{N}} L_{cls} (\hat{p}_i,\hat{p}_i^\star)
+ \frac{1}{\hat{N}_{+}} \sum_{i=1}^{\hat{N}_+} L_{reg} (\hat{\mathbf{d}}_i,\hat{\mathbf{d}}_i^\star)  \notag \\
&+ \frac{1}{\hat{N}_{+}} \sum_{i=1}^{\hat{N}_{+}} L_{rec} (\mathbf{Y}^{(i)}, \mathbf{s}^{(i)})
\end{align}
where $\hat{N} \leq 512$ is the number of text proposals sampled after hard negative mining,
and $\hat{N}_{+} \leq 256$ is the number of positive ones.
The thresholds for positive and negative anchors are set to $0.6$ and $0.4$ respectively, {\color{black}{so as to increase the difficulty for text classification and regression, and improve the ability of TDN.}}
$\hat{p}_i$ and $\hat{\mathbf{d}}_i$ are the outputs of TDN.
$\mathbf{s}^{(i)}= \{ \mathbf{s}^{(i)}_1, \dots, \mathbf{s}^{(i)}_{T+1} \}$  is the ground-truth tokens for sample $i$, where $\mathbf{s}^{(i)}_{T+1}$ represents the special ``END'' token,  and $\mathbf{Y}^{(i)} = \{ \mathbf{y}^{(i)}_1, \dots, \mathbf{y}^{(i)}_{T+1} \}$
is the corresponding output sequence of decoder LSTMs.
$L_{rec}(\mathbf{Y}, \mathbf{s}) = - \sum_{t=1}^{T+1} \log \mathbf{y}_t(s_{t})$ denotes the cross entropy loss on $\mathbf{y}_1, \dots, \mathbf{y}_{T+1}$,
where $\mathbf{y}_t(s_{t})$ represents the predicted probability of the output being $s_t$ at
time-step $t$.

\section{Experiments}
In this section, we
conduct
extensive experiments to verify the effectiveness of the proposed method. We first describe {\color{black}{the used}} datasets and the implementation details. Then, our model is compared
against
a few
state-of-the-art
methods
on a number of standard benchmark datasets, including both regular and irregular text in natural scene images.
Intermediate results are also demonstrated for ablation study.
Finally, we compare the difference between the new model and the ones
presented in our previous conference versions.

\subsection{Datasets}
\label{dataset}
The following datasets are used in our experiments for training and evaluation:

\noindent{\bf Synthetic Datasets} (\textbf{SynT}) \quad
In~\cite{Gupta16}, a fast and scalable engine was presented to generate synthetic images of text in clutter. A synthetic dataset with $800,000$ images (denoted as ``SynthText'') was also released to public. It contains a large number of multi-oriented text instances, and is adopted widely in model pre-training.

\noindent{\bf ICDAR2013} (\textbf{IC13})~\cite{icdar2013}  \quad
This is the widely used dataset for scene text spotting from ICDAR2013 Robust Reading Competition.
Images in this dataset explicitly focus around the text content of interest, which results in well-captured, nearly horizontal text instances. There are $229$ images for training and $233$ images for test. Text instances are annotated by horizontal bounding boxes with word-level transcriptions. There  are $3$ specific lists of words provided as lexicons for reference in the test phase, i.e., ``Strong'', ``Weak'' and ``Generic''. ``Strong'' lexicon provides $100$ words per-image including all words appeared in the image. ``Weak'' lexicon contains all words appeared in the entire dataset, and ``Generic'' lexicon is a $90$k word vocabulary proposed by~\cite{Max2016IJCV}.

\noindent{\bf ICDAR02015} (\textbf{IC15})~\cite{icdar2015}   \quad
This is another popular dataset from ``Incidental Scene Text'' of ICDAR2015 Robust Reading Competition. Images in this dataset are captured incidentally with Google Glasses, and hence most text instances are irregular (oriented, perspective and blurring). There are $1,000$ images for training and $500$ images for test. $3$ scales of lexicons are also provided in test phase. The ground-truth for text is given by quadrangles and word-level annotations.

\noindent{\bf Total-Text} (\textbf{TT})~\cite{totaltext}   \quad
This dataset was released in ICDAR2017, featuring curved-oriented text. More than half of its images have a combination of text instances with more than two orientations. There are $1,255$ images in training set and $300$ images for test. Text is annotated by polygon at the word level.

\noindent{\bf MLT}~\cite{MLT} \quad
MLT is a large multi-lingual text dataset, which contains $7,200$ training images, $1,800$ validation images and $9,000$ test images. As adopted by FOTS~\cite{FOTS2018}, we also employ the ``Latin'' instances in training and validation images to enlarge our training data. Because our proposed model is only for reading English words, we cannot test the model on MLT test dataset.

\noindent{\bf AddF2k}~\cite{Zhong2016} \quad
It contains $1,715$ images with near horizontal text instances released in~\cite{Zhong2016}. The images are annotated by horizontal bounding boxes and word-level transcripts. All images are used in training phase.

\noindent{\bf COCO-Text} (\textbf{CT})~\cite{cocotext}   \quad
COCO-Text is %
by far
the largest dataset for scene text detection and recognition. It consists of $43,686$ images for training, $10,000$ images for validation and another $10,000$ for test. In our experiment, we collect all training and validation images for training. COCO-Text is created by annotating images from the MS COCO dataset, which contains images of complex scenes. As a result, this dataset is very challenging with text in arbitrary shapes. The ground-truth is given by word-level with top-left and bottom-right coordinates. {\color{black}{Note that images in this dataset are only used to fine-tune the model when evaluating on COCO-Text test data.}}

\begin{table*}[!ht]
	\newcommand{\tabincell}[2]{\begin{tabular}{@{}#1@{}}#2\end{tabular}}
	\begin{center}
		\caption{Text spotting results on ICDAR2013 dataset.  We present F-measures here in percentage. Using ResNet$50$ as backbone, Our model achieves state-of-the-art performance on ``Word-Spotting''. The approaches marked with ``*'' need to be trained with additional character-level annotations. In each column, the best result is shown in $\textbf{bold}$ font, and the second best is shown in $\mathit{italic}$ font.}
		\label{Tab:ic13}
		\scalebox{0.95}{
			\begin{tabular}{l|c|c|c|c|c|c|c|c}
				\hline
				Method & Backbone & \tabincell{c}{Training \\ data}& \multicolumn{3}{|c}{\tabincell{c}{ICDAR2013 \\ Word-Spotting}} &  \multicolumn{3}{|c} {\tabincell{c}{ICDAR2013 \\ End-to-End}}  \\\cline{4-9} 	&&& \multicolumn{1}{c|}{Strong} & \multicolumn{1}{|c|}{Weak}  & \multicolumn{1}{|c|}{Generic}   & \multicolumn{1}{c|}{Strong} & \multicolumn{1}{|c|}{Weak} & \multicolumn{1}{|c}{Generic}  \\
				\hline
				Deep2Text II+~\cite{Yin2014pami} & - & IC13 & $84.84$ & $83.43$ & $78.90$ & $81.81$ & $79.47$ & $76.99$ \\
				\hline
				Jaderberg~\etal~\cite{Max2016IJCV} & CNN & SynT+IC13+IC03+SVT & $90.49$ & $-$ & $76$ & $86.35$ & $-$ & $-$  \\
				\hline
				FCRNall+multi-filt~\cite{Gupta16} & VGG16 & SynT & $-$ & $-$ & $84.7$ & $-$ & $-$ & $-$  \\
				\hline
				TextBoxes~\cite{LiaoAAAi2017} & VGG16 & SynT+IC13  & $93.90$ & $91.95$ & $85.92$ & $91.57$ & $89.65$ & $83.89$ \\
				\hline
				DeepTextSpotter~\cite{Textspotter_ICCV2017} & GoogleNet & SynT+IC13+IC15   & $92$ & $89$ & $81$ & $89$ & $86$ & $77$  \\
				\hline
				TextBoxes++~\cite{Liao2018Text} & VGG16+CRNN & SynT+IC13  & $95.50$ & $\mathit{94.79}$ & $87.21$ & $\mathbf{92.99}$ & $\mathbf{92.16}$ & $84.65$  \\
				\hline
				MaskTextSpotter*~\cite{Maskspot2018} & R50 & SynT+IC13+IC15+TT   & $92.5$ & $92.0$ & $\mathit{88.2}$ & $92.2$ & $91.1$ & $\mathbf{86.5}$  \\
				\hline
				TextNet~\cite{Sun2018TextNetIT} & R50 & SynT+IC13   & $94.59$ & $93.48$ & $86.99$ & $89.77$ & $88.80$ & $82.96$  \\
				\hline
				AlignmentTextSpotter*~\cite{hetong2018} & PVA & SynT+IC13+IC15+MLT   & $93$ & $92$ & $87$ & $91$ & $89$ & $\mathit{86}$  \\
				\hline
				FOTS~\cite{FOTS2018} & R50 & SynT+IC13+MLT  & $\mathit{95.94}$ & $93.90$ & ${87.76}$ & $91.99$ & $90.11$ & $84.77$  \\
				\hline
				\hline
				Ours & R50 & SynT+IC13+IC15+MLT  & $\mathbf{96.39}$ & $\mathbf{95.53}$ & $\mathbf{89.45}$  & $\mathit{92.56}$ & $\mathit{91.60}$ & $85.49$  \\
				\hline
				\end{tabular}
		}
	\end{center}
\end{table*}

\begin{table*}[!t]
	\
	\newcommand{\tabincell}[2]{\begin{tabular}{@{}#1@{}}#2\end{tabular}}
	\begin{center}
		\caption{Text spotting results on ICDAR2015 dataset. We present  F-measures here in percentage.
			Our model achieves promising performance on both ``Word-Spotting'' and ``End-to-End'' settings, in comparison with other methods. The approaches marked with ``*'' need to be trained with additional character-level annotations. In each column, the best performing result is shown in $\textbf{bold}$ font, and the second best is shown in $\mathit{italic}$ font.}
		\label{Tab:ic15}
		\scalebox{0.95}{
			\begin{tabular}{l|c|c|c|c|c|c|c|c}
				\hline
				Method & Backbone & \tabincell{c}{Training \\ data}& \multicolumn{3}{|c}{\tabincell{c}{ICDAR2015 \\ Word-Spotting}} &  \multicolumn{3}{|c} {\tabincell{c}{ICDAR2015 \\ End-to-End}}  \\\cline{4-9} 	&&	& \multicolumn{1}{c|}{Strong} & \multicolumn{1}{|c|}{Weak}  & \multicolumn{1}{|c|}{Generic}   & \multicolumn{1}{c|}{Strong} & \multicolumn{1}{|c|}{Weak} & \multicolumn{1}{|c}{Generic}  \\
				\hline
				Deep2Text-MO~\cite{Yin2014pami} & - & IC11 & $17.58$ & $17.58$ & $17.58$ & $16.77$ & $16.77$ & $16.77$ \\
				\hline
				TextSpotter~\cite{textspotter} & - & IC15 & $-$ & $-$ & $-$ & $35.0$ & $19.9$ & $15.6$  \\
				\hline
				TextProposals + DictNet~\cite{TextProposals,maxNIPS14} & CNN & SynT & $56.00$ & $52.26$ & $49.73$ & $53.30$ & $49.61$ & $47.18$  \\
				\hline
				DeepTextSpotter~\cite{Textspotter_ICCV2017} & GoogleNet & SynT+IC13+IC15 & $58$ & $53$ & $51$ & $54$ & $51$ & $47$  \\
				\hline
				TextBoxes++~\cite{Liao2018Text} & VGG16+CRNN & SynT+IC15 & $76.45$ & $69.04$ & $54.37$ & $73.34$ & $65.87$ & $51.90$  \\
				\hline
				ASTER~\cite{shiPAMI2018} & VGG16+R50 & SynT+IC15 & $75.2$ & $71.3$ & $67.6$ & $70.6$ & $67.3$ & $64.0$  \\
				\hline
				MaskTextSpotter*~\cite{Maskspot2018} & R50 & SynT+IC13+IC15+TT  & $79.3$ & $74.5$ & $64.2$ & $79.3$ & $73.0$ & $62.4$  \\
				\hline
				TextNet~\cite{Sun2018TextNetIT} & R50 & SynT+IC15  & $82.38$ & $78.43$ & $62.36$ & $78.66$ & $74.90$ & $60.45$  \\
				\hline
				AlignmentTextSpotter*~\cite{hetong2018} & PVA & SynT+IC13+IC15+MLT  & $85$ & $80$ & $65$ & $82$ & $77$ & $63$  \\
				\hline
				FOTS~\cite{FOTS2018} & R50 & SynT+IC13+IC15+MLT  & $\mathit{87.01}$ & $\mathbf{82.39}$ & ${67.97}$ & $\mathit{83.55}$ & $\mathbf{79.11}$ & $\mathit{65.33}$  \\
				\hline
				TextDragon~\cite{textdragon} & VGG16 & SynT+IC15  & $86.22$ & $ \mathit{81.62}$ & $\mathit{68.03}$ & $82.54$ & $\mathit{78.34}$ & $65.15$  \\

				\hline
				\hline
			 	Ours  & R50 & SynT+IC13+IC15+MLT   & $\mathbf{87.80}$ & $81.58$ & $\mathbf{68.21}$  & $\mathbf{84.23}$ & $78.04$ & $\mathbf{65.53}$  \\
				\hline
			\end{tabular}
		}
	\end{center}
\end{table*}

\begin{figure*}[htbp]
	\begin{center}
		\includegraphics[width=0.8\textwidth]{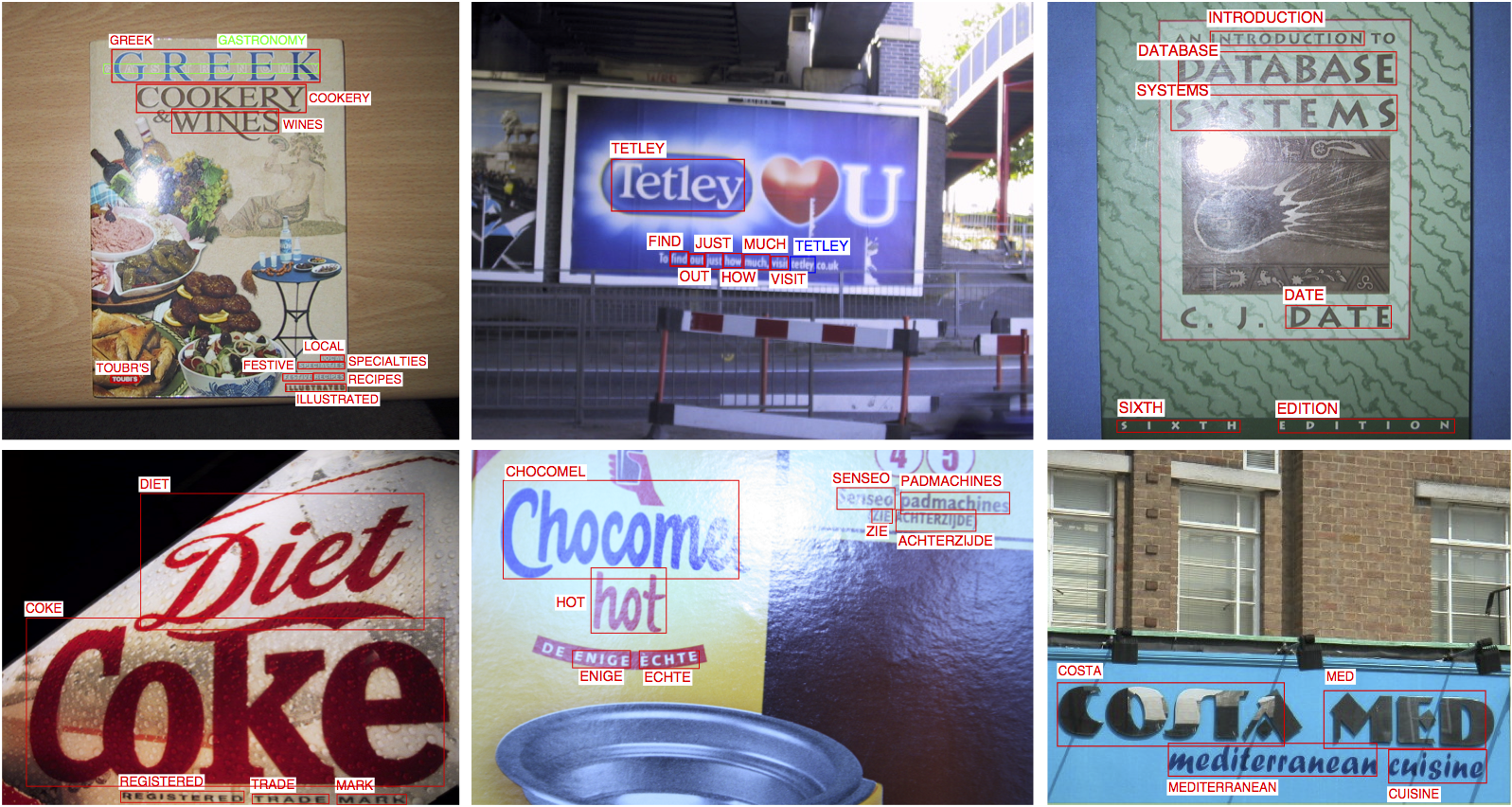}
	\end{center}
	\caption{Examples of text spotting results on ICDAR2013. The red bounding boxes are both detected and recognized correctly. The green bounding boxes are missed words. The new model can cover more scales of text compared to the conference version~\cite{li2017towards}. For example, ``SIXTH'' and ``EDITION'' in the third image can be covered, which have a big space between characters.
	}
	\label{fig:ic13res}
\end{figure*}

\begin{figure*}[htbp]
	\begin{center}
		\includegraphics[width=0.8\textwidth]{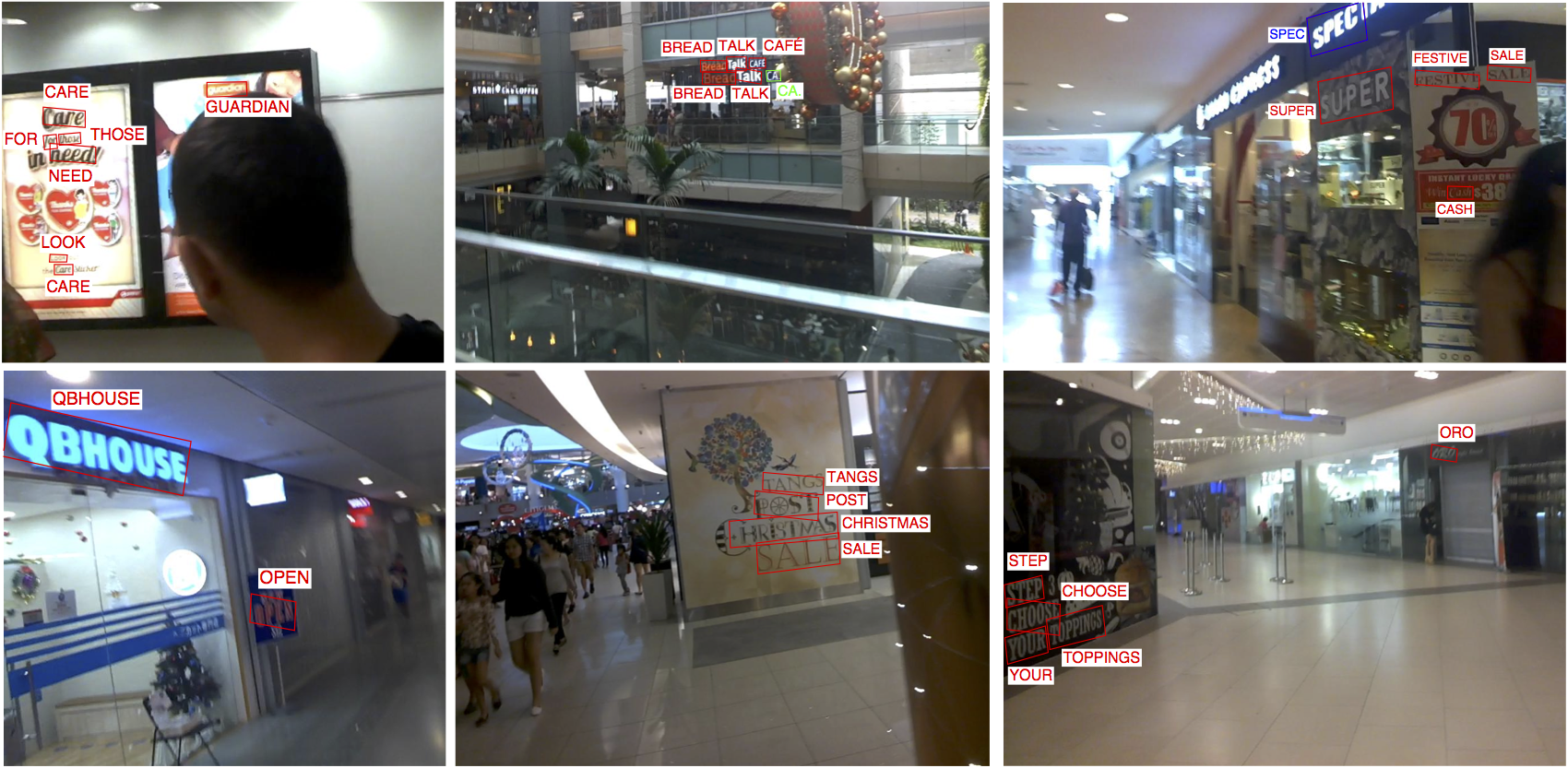}
	\end{center}
	\caption{Examples of text spotting results on ICDAR2015. The red bounding boxes are both detected and recognized correctly. The green bounding boxes are missed words, and the blue labels are wrongly recognized. With the employed $2$D attention mechanism, our network is able to detect and recognize oriented text with a single forward pass in cluttered natural scene images.
	}
	\label{fig:ic15res}
\end{figure*}

\subsection{Implementation Details}

In contrast to the work in our conference version~\cite{li2017towards} where the network is trained with TRN module locked initially, in this work, we train the whole network in an end-to-end fashion during the entire training process. This is achieved, we believe, with the benefit of better text proposals and RoI-Align methods. We use an approximate joint training process~\cite{renNIPS15fasterrcnn} to minimize the aforementioned two losses, i.e., $L_{T\!P\!N}$ and $L_{D\!R\!N}$ together, ignoring the derivatives with respect to the proposed boxes' coordinates.

The whole network is trained end-to-end on ``SynthText'' for $2$ epochs firstly. Then we randomly sample $10$k images from ``SynthText'', and combine with real training data to fine-tune the model for $20$ epochs. Lastly, synthetic data is removed and the model is fine-tuned using only real data for another $15$ epochs. {\color{blue}{Different real training datasets are %
used
for different tasks, which will be discussed %
in the experiments.
}}

We optimize our model using SGD with a batch size of $4$, a weight decay of $0.0001$ and a momentum of $0.9$. The learning rate is set to $0.005$ initially, with a decay rate of $0.8$ every $30$k iterations until it reaches $10^{-4}$ on the synthetic training data. When fine-tuning on real training images, the learning rate is decayed again with a rate of $0.8$ every $30$k iterations until it reaches $10^{-5}$.

Data augmentation is also
employed
in the model training process.
Specifically,
1) A multi-scale training strategy is used, where the shorter side of input image is randomly resized to three scales of $(600, 800, 1000)$ pixels, and the longer side is no more than $1200$ pixels.
2) We randomly re-scale (with a probability of $0.5$) the height of the image with a ratio from $0.8$ to $1.2$ without changing its width, so that the bounding boxes have more variable aspect ratios.
{\color{blue}
{3) Images are rotated in the range of
[$-10^\circ$, $10^\circ$]
randomly with a probability of $0.4$. 4) Images are randomly cropped from the input with a proportion of $0.9$ and then resized to the original size. }}

During the test phase, we re-scale the input image into multiple sizes as well so as to cover the large range of bounding box scales. At each scale, $300$ proposals with the highest textness scores are produced by TPN. Those proposals are re-identified by TDN and recognized by TRN simultaneously. A recognition score is then calculated by averaging the output probabilities. The ones with textness score larger than $0.5$ and recognition score larger than $0.7$ are kept and merged via NMS (non-maximum suppression) as the final output.

{\color{black}{\subsection{Evaluation Criterion}}}

We follow the standard evaluation criterion in the end-to-end text spotting task: a bounding box is considered as correct if its IoU ratio with any ground-truth is greater than $0.5$ and the recognized word also matches, ignoring the case. The ones with no longer than three characters and annotated as ``do not care'' are ignored. For the ICDAR2013 and ICDAR2015 datasets,
there are two protocols: ``End-to-End" and ``Word Spotting". ``End-to-End" protocol requires all words in the image to be recognized, %
no matter
whether the string exists or not in the provided contextualised lexicon.
``Word Spotting" on the other hand, only looks at the words that actually exist in the lexicon provided, ignoring all the rest that do not appear in the lexicon. There is no lexicon released in COCO-Text and Total-Text. Thus methods {\color{black}{will be}} evaluated based on raw outputs, without using any prior knowledge.
It should be noted that the location ground-truth is \textit{rectangles} in ICDAR2013 and COCO-Text, \textit{quadrangles} in ICDAR2015, and \textit{polygons} in Total-Text.

\begin{table*}[t!]
	\newcommand{\tabincell}[2]{\begin{tabular}{@{}#1@{}}#2\end{tabular}}
	\begin{center}
		\caption{Text detection and text spotting results on Total-Text dataset. ``Ours (New\_R$101$)'' achieves the best ``End-to-End'' performance among the compared methods.  In each column, the best result is shown in $\textbf{bold}$ font, and the second best result is shown in $\mathit{italic}$ font.}
		\label{Tab:total}
		\scalebox{0.95}{
			\begin{tabular}{l|c|c|c|c|c|c}
				\hline
				Method & Backbone & \tabincell{c}{Training \\ data}& \multicolumn{3}{|c}{\tabincell{c}{Detection}} &  \multicolumn{1}{|c} {\tabincell{c}{ End-to-End}}  \\\cline{4-7} &&	& \multicolumn{1}{c|}{Recall} & \multicolumn{1}{|c|}{Precision}  & \multicolumn{1}{|c|}{F-measure}   &  \multicolumn{1}{c}{F-measure}  \\
				\hline
				DeconvNet~\cite{totaltext}  & VGG16 & SynT+IC13+IC15+MLT  & $33.0$ & $40.0$ & $36.0$ & $-$ \\
				\hline
				TextBoxes~\cite{LiaoAAAi2017} & VGG16 & SynT+IC13 & $45.5$ & $62.1$ & $52.5$ & $36.3$ \\
				\hline
			  	MaskTextSpotter~\cite{Maskspot2018}  & R50 & SynT+IC13+IC15+TT & $55.0$ & $\mathit{69.0}$ & $61.3$ & $52.9$ \\
				\hline
				TextNet~\cite{Sun2018TextNetIT}  & R50 & SynT+TT & ${59.45}$ & ${68.21}$ & $\mathit{63.53}$ & ${54.02}$ \\
				\hline
				TextDragon~\cite{textdragon}  & R50 & SynT+TT & $\mathbf{75.7}$ & $\mathbf{85.6}$ & $\mathbf{80.3}$ & $48.8$ \\
				\hline
				\hline
			Ours  & R50 & SynT+IC13+IC15+TT & $59.38$ & $63.25$ & $61.25$  & $\mathbf{58.56}$  \\

				\hline
			\end{tabular}
		}
	\end{center}
\end{table*}

\begin{table*}[htbp]
	\newcommand{\tabincell}[2]{\begin{tabular}{@{}#1@{}}#2\end{tabular}}
	\begin{center}
		\caption{Text detection and text spotting results on COCO-Text dataset. Our method achieves state-of-the-art text detection performance, with F-measure outperforming the second best around $4 \%$. The end-to-end performance is also promising. }
		\label{Tab:coco}
		\scalebox{0.95}{
			\begin{tabular}{l|c|c|c|c|c|c}
				\hline
				Method & Backbone & \tabincell{c}{Training \\ data}& \multicolumn{3}{|c}{\tabincell{c}{Detection}} &  \multicolumn{1}{|c} {\tabincell{c}{ End-to-End}}  \\\cline{4-7} &&	& \multicolumn{1}{c|}{Recall} & \multicolumn{1}{|c|}{Precision}  & \multicolumn{1}{|c|}{F-measure}   &  \multicolumn{1}{c}{\tabincell{c}{Average \\ Precision}}  \\
				\hline
				Yao~\etal~\cite{yao_holistic} & VGG16 & IC13+IC15+MSRA-TD500 & $27.10$ & $43.23$ & $33.31$ & $-$ \\
				\hline
				He~\etal~\cite{He2017ICCV} & VGG16 & IC13+IC15+ext.  & $31$ & $46$ & $37$ & $-$ \\
				\hline
				EAST~\cite{EAST17} & VGG16 & ImageNet+CT & $32.40$ & $50.39$ & $39.45$ & $-$ \\
				\hline
				TO-CNN~\cite{Prasad2018}& VGG16 & NTU-UTOI  & $44$ & $47$ & $45$ & $-$ \\
				\hline
				TextBoxes++~\cite{Liao2018Text} & VGG16 & SynT+CT & $56.70$ & $60.87$ & $58.72$ & $-$ \\
				\hline
				Lyu~\etal~\cite{Lyucvpr18} & VGG16 & SynT+IC15 & $52.9$ & $72.5$ & $61.1$ & $-$ \\
				\hline
				MaskTextSpotter+~\cite{LiaoPAMI2019} & R50 & SynT+IC13+IC15+TT+AddF2k & $\mathbf{58.3}$ & $66.8$ & $62.3$ & $23.9$ \\
				\hline
				\hline
				Ours & R50 & SynT+IC13+IC15+MLT+CT & ${56.60}$ & $\mathbf{74.76}$ & $\mathbf{64.43}$  & $\mathbf{33.75}$   \\
				\hline
			\end{tabular}
		}
	\end{center}
\end{table*}

\vspace{4mm}
{\color{black}{\subsection{Comparison with State-of-the-art Methods}}}

\subsubsection{\bf Experimental Results on ICDAR2013}

{\color{black}{The ICDAR2013 dataset is mostly used to evaluate model ability in detecting and recognizing horizontal text. {\color{blue}{We use training images from IC13, IC15 and MLT during fine-tuning process.}} The text spotting results  are presented and compared with other state-of-the-art methods in Table~\ref{Tab:ic13}. Using ResNet$50$ as backbone, our model outperforms existing methods under ``Word-Spotting'' protocol, with about $1.2 \%$ F-measure improvement when using generic lexicon. %
Results under ``End-to-End'' protocol are also comparable with other methods.
Samples of results on ICDAR2013 are visualized in Figure~\ref{fig:ic13res}.
}}
\subsubsection{\bf Experimental Results on ICDAR2015}

We verify the effectiveness of the proposed model in spotting oriented text on the ICDAR2015 dataset.
{\color{blue}{Real training images from IC13, IC15 and MLT are adopted during the fine-tuning process.}}
As shown in Table~\ref{Tab:ic15}, our model achieves state-of-the-art performance under two task settings with both protocols (excluding the ``weak'' one). Note that we have not used any lexicon in the ``Generic'' sub-task. The results are the raw outputs without using any prior knowledge about lexicon. %
Some qualitative results are presented in Figure~\ref{fig:ic15res}, with both quadrangle localizations and corresponding text labels. It can be seen that with the help of the spatial $2$D attention weights, the improved framework is able to tackle irregular text.

\begin{figure*}[!ht]
	\begin{center}
		\includegraphics[width=0.8\textwidth]{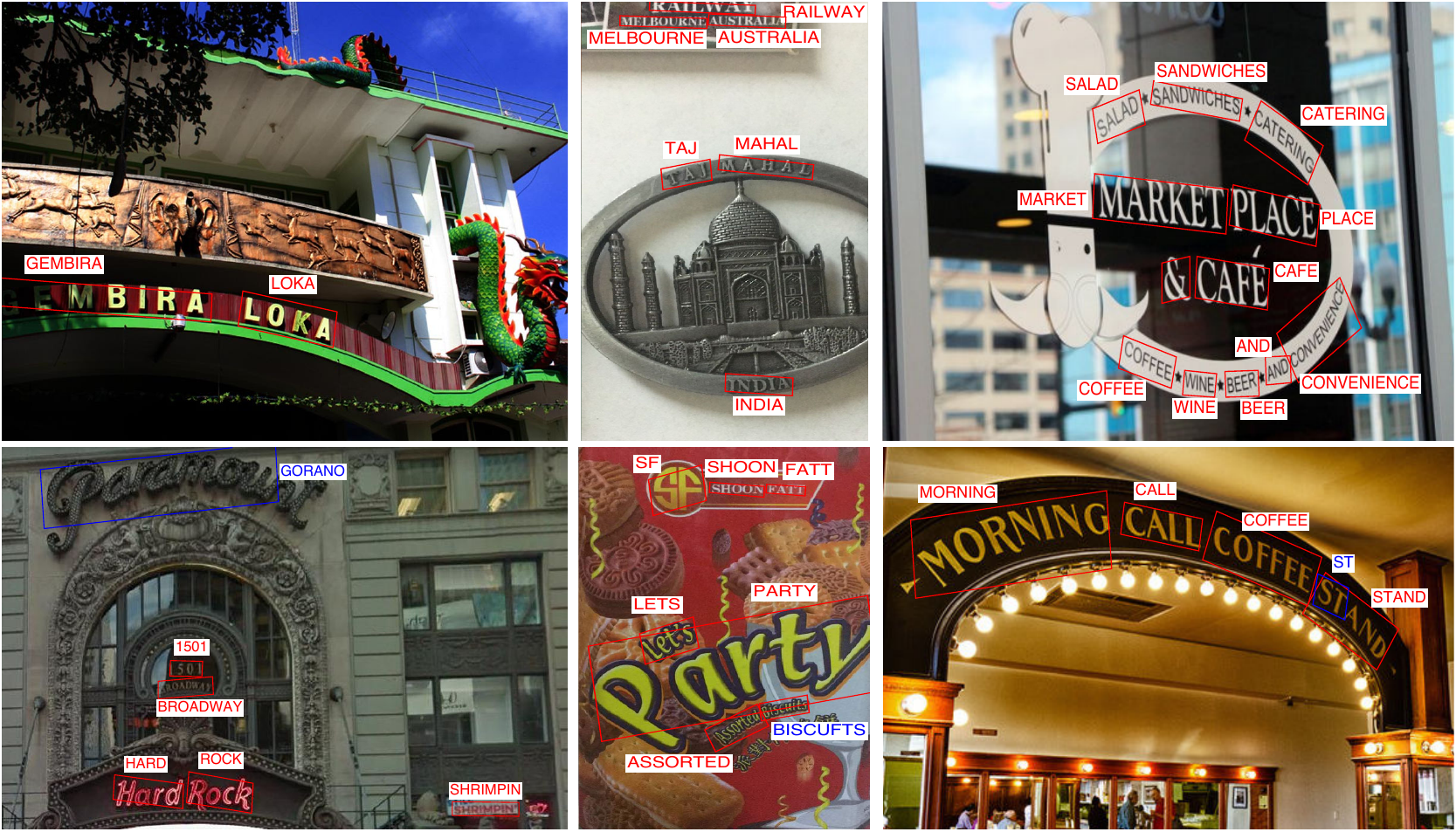}
	\end{center}
	\caption{Text spotting examples on Total-Text. The red bounding boxes are both detected and recognized correctly. The blue ones are recognized incorrectly. The use of $2$D attention mechanism enables our model detect and recognize curved text with a single forward pass in cluttered natural scene images.
	}
	\label{fig:totalres}
\end{figure*}

\begin{figure*}[!ht]
	\begin{center}
		\includegraphics[width=0.8\textwidth]{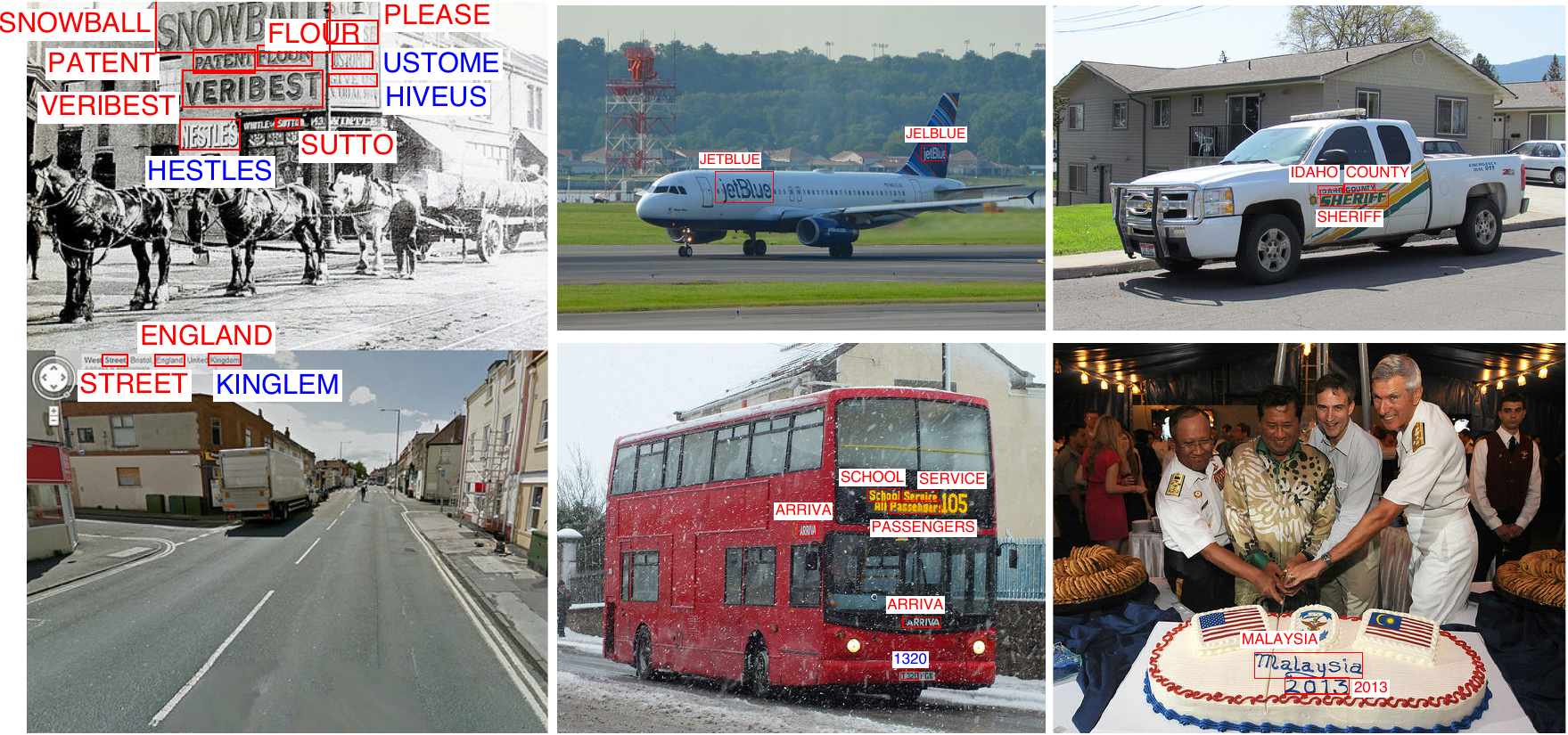}
	\end{center}
	\caption{Text spotting examples on COCO-Text. The red bounding boxes are both detected and recognized correctly. The blue labels are wrongly recognized.
	}
	\label{fig:cocores}
\end{figure*}

\subsubsection{\bf Experimental Results on Total-Text}

Next, we conduct experiments on the Total-Text dataset to evaluate the performance of our method for curved text. {\color{blue}{Real training images from IC13, IC15 and Total-Text are employed here %
for fine-tuning.%
}}
As shown in Table~\ref{Tab:total}, based on the same evaluation protocol as that used in~\cite{Maskspot2018} where the IoU is calculated by {\color{black}{using}} polygon ground-truth bounding boxes, our model leads to an ``End-to-End'' performance of $58.56 \%$ without using any lexicon, which is about $4.5\%$ higher than the best of the compared methods. %

{\color{black}{Text detection results on Total-Text are also presented here for reference. As our model is not delicately designed for text detection, the output bounding box does not enclose the text tightly, which leads to unsatisfactory detection performance under the detection evaluation criterion. However, the promising end-to-end performance exactly proves the strength and robustness of our 2D attention based text recognition model from another point of view. Compared to TextDragon~\cite{textdragon} that leads to the best detection performance, its end-to-end result is about $10 \%$ lower than ours. Our method can correctly recognize text contained in loose bounding boxes. That is of practical importance from the viewpoint of text spotting.}} Some visualization results are presented in Figure~\ref{fig:totalres}.

\subsubsection{\bf Experimental Results on COCO-Text}

The COCO-text dataset %
is very challenging, not only because of the quantity, but also lying in the large variance of text appearance. %
{\color{blue}{Here, we fine-tune the pre-trained model with real training images from IC13, IC15, MLT and COCO-Text.}}
The corresponding experimental results are shown in Table~\ref{Tab:coco}.
{\color{black}{It should be noted that text ground-truth in COCO-Text is given by rectangles. With this setting, our model achieves the highest text detection performance, which is $2 \%$ higher than the state-of-the-art.
With the further integration of a strong text recognizer, our model finally achieves a surpassing text spotting performance.
Several text spotting examples on COCO-Text are visualized in Figure~\ref{fig:cocores}.}}

\begin{table*}[!ht]
	\newcommand{\tabincell}[2]{\begin{tabular}{@{}#1@{}}#2\end{tabular}}
	\begin{center}
		\caption{\color{black}{Experiments on multi-task learning. We present F-measures here in percentage. ``Former'' ones show results from previous conference version~\cite{li2017towards}. Experimental results from both the conference version~\cite{li2017towards} and
		our
		proposed
		model proves that the joint training of the whole framework benefits the end-to-end task. }}
		\label{Rebt:multi-task}
		\scalebox{0.8}{
			\begin{tabular}{l|c|c|c|c|c|c|c|c|c|c|c|c|c|c|c|c}
				\hline
				{\tabincell{c}{Model \\ Name}} & {\tabincell{c}{Backbone}}  & Attn. & Training   & \multicolumn{3}{|c}{\tabincell{c}{ICDAR2013 \\ Word-Spotting}} &  \multicolumn{3}{|c} {\tabincell{c}{ICDAR2013 \\ End-to-End}} &
				\multicolumn{3}{|c}{\tabincell{c}{ICDAR2015 \\ Word-Spotting}} &  \multicolumn{3}{|c|} {\tabincell{c}{ICDAR2015 \\ End-to-End}} & Total-Text
				 \\\cline{5-16} &&&&
				\multicolumn{1}{c|}{Strong} & \multicolumn{1}{|c|}{Weak}  & \multicolumn{1}{|c|}{Generic}   & \multicolumn{1}{c|}{Strong} & \multicolumn{1}{|c|}{Weak} & \multicolumn{1}{|c|}{Generic} 	&			\multicolumn{1}{c|}{Strong} & \multicolumn{1}{|c|}{Weak}  & \multicolumn{1}{|c|}{Generic}   & \multicolumn{1}{c|}{Strong} & \multicolumn{1}{|c|}{Weak} & \multicolumn{1}{|c|}{Generic} \\
				\hline
			{\tabincell{c}{Former (sep)~\cite{li2017towards} }} & VGG-16   & $1$D  & Separate   &  $92.94 $ & $90.54 $ & $84.24$  & $88.20$ & $86.06$ & $81.97$ & - & - & -  & - & - & - & -  \\
			\hline
			{\tabincell{c}{Former (full)~\cite{li2017towards}}}  & VGG-16  & $1$D & Joint  &   $94.16$ & $92.42$ & $88.20$  & $91.08$ & $89.81$ & $84.59$ & - & - & -  & - & - & - & -  \\
			\hline
	 		{\tabincell{c}{Ours (sep)}}  & R$50$+FPN   & $2$D & Separate   & $95.95$ & $94.31$ & $88.35$  & $91.28$ & $90.90$ & $84.35$ & $82.67$ & $77.27$ & $63.82$  & $79.86$ & $75.12$ & $61.04$ & $57.82$ \\
			\hline
	 		Ours & R$50$+FPN  & $2$D & Joint   & $96.35$ & $94.87$ & $88.90$  & $92.13$ & $91.25$ & $84.74$ & $85.64$ & $80.45$ & $65.84$  & $82.21$ & $77.14$ & $63.55$ & $58.72$ \\
			\hline
			\end{tabular}
		}
	\end{center}
\end{table*}

\begin{table*}[htbp]
	\newcommand{\tabincell}[2]{\begin{tabular}{@{}#1@{}}#2\end{tabular}}
	\begin{center}
		\caption{\color{black}{Text detection results on ICDAR2013, ICDAR2015 and Total-Text. F-measures are presented here in percentage. We use standard detection evaluation criteria proposed in each dataset for fair comparison. The %
		inclusion
		of the recognition loss in model training greatly enhances the detection performance.}}
		\label{Tab:det}
		\scalebox{0.92085}{
			\begin{tabular}{l|c|c|c|c|c}
				\hline
			    Model Name    & \multicolumn{3}{|c|}{ICDAR2013} &
				\multicolumn{1}{|c|}{ICDAR2015}  & Total-Text
				 \\\cline{2-4} &
				\multicolumn{1}{c|}{ICDAR standard} & \multicolumn{1}{|c|}{DetEval}  & \multicolumn{1}{|c|}{IoU} &  \\
				\hline
				Jaderberg~\etal~\cite{Max2016IJCV} & -  & - & $76.2$ & - & -  \\
				\hline
				FCRNall+multi-filt~\cite{Gupta16} & -  & - & $84.2$ & - & -   \\
				\hline
				CTPN~\cite{Tian2016} & $82.2$ & $87.7$ & - & $61$ & - \\
				\hline
				TextBoxes~\cite{LiaoAAAi2017} & $85$ & $86$ & - & - & $52.5$ \\
				\hline
				SSTD~\cite{He2017ICCV} & $87$ & $88$ & - & $77$ & - \\
				\hline
				RRPN~\cite{RRPN} & $-$ & $91$ & - & $80.2$ & - \\
				\hline
				AlignmentTextSpotter\_Det~\cite{hetong2018} & $88$ & $88$ & - & $83$ & - \\
				\hline
				AlignmentTextSpotter~\cite{hetong2018} & $90$ & $90$ & - & $87$ & - \\
				\hline
				FOTS\_Det~\cite{FOTS2018} & $86.9$ & $87.3$ & - & $85.3$ & - \\
				\hline
				FOTS~\cite{FOTS2018}  & $92.5$ & $92.8$ & - & $89.8$ & - \\
				\hline
				MaskTextSpotter~\cite{Maskspot2018}   & - & $91.7$ & - & $86.0$ & $61.3$ \\
				\hline
				TextNet~\cite{Sun2018TextNetIT}   & $91.3$ & $91.4$ & - & $87.4$ & $63.5$ \\
				\hline
				\hline
				\tabincell{c}{Former (sep)~\cite{li2017towards}}  & -  & - & $83.4$ & - & -  \\
				\hline
				\tabincell{c}{Former (full)~\cite{li2017towards}}  & -  & - & $85.6$  & - & - \\
				\hline
				\tabincell{c}{Ours (sep)}  & $87.7$  & $88.5$ & $87.5$ & $79.5$ & $60.1$  \\
				\hline
				\tabincell{c}{Ours}  & $91.6$  & $92.3$ & $90.3$  & $85.0$ & $61.5$ \\
				\hline
			\end{tabular}
		}
	\end{center}
\end{table*}

\begin{table*}[!ht]
	\newcommand{\tabincell}[2]{\begin{tabular}{@{}#1@{}}#2\end{tabular}}
	\begin{center}
	\caption{\color{black}{Ablation experiments on RoI pooling manner. ``Former'' ones show results from the conference version~\cite{li2017towards}. ``RoI'' column means using \textbf{F}ixed-size or \textbf{V}arying-size RoI pooling. ``RoI Spec.'' shows the RoI feature size specification, where the maximum size is presented for \textbf{V}arying-size RoI pooling. The proposed varying-size RoI pooling is more flexible in handling the large variability on text scales and aspect ratios, and leads to the best performance. Note that we use $W_{max}=35$ in~\cite{li2017towards} but $30$ here, which leads to a faster running speed but without sacrificing model performance. }}
		\label{Tab:roi}
		\scalebox{0.78}{
			\begin{tabular}{l|c|c|c|c|c|c|c|c|c|c|c|c|c|c|c|c|c}
				\hline
				{\tabincell{c}{Model \\ Name}} & {\tabincell{c}{Backbone }}  & Attn. & {\tabincell{c}{RoI }} & {\tabincell{c}{RoI \\ Spec.}}  & \multicolumn{3}{|c}{\tabincell{c}{ICDAR2013 \\ Word-Spotting}} &  \multicolumn{3}{|c} {\tabincell{c}{ICDAR2013 \\ End-to-End}} &
				\multicolumn{3}{|c}{\tabincell{c}{ICDAR2015 \\ Word-Spotting}} &  \multicolumn{3}{|c|} {\tabincell{c}{ICDAR2015 \\ End-to-End}} & Total-Text
				 \\\cline{6-17} &&&&&
				\multicolumn{1}{c|}{Strong} & \multicolumn{1}{|c|}{Weak}  & \multicolumn{1}{|c|}{Generic}   & \multicolumn{1}{c|}{Strong} & \multicolumn{1}{|c|}{Weak} & \multicolumn{1}{|c|}{Generic} 	&			\multicolumn{1}{c|}{Strong} & \multicolumn{1}{|c|}{Weak}  & \multicolumn{1}{|c|}{Generic}   & \multicolumn{1}{c|}{Strong} & \multicolumn{1}{|c|}{Weak} & \multicolumn{1}{|c|}{Generic} \\
				\hline
			Former (fix)~\cite{li2017towards} & VGG-16   & $1$D  & F & $4 \times 20$   & $93.33 $ & $91.66 $ & $87.73$  & $90.72 $ & $87.86$ & $83.98 $ & - & - & -  & - & - & - & -  \\
			\hline
			Former (full)~\cite{li2017towards} & VGG-16  & $1$D &  V & $4 \times 35$   &   $94.16$ & $92.42$ & $88.20$  & $91.08$ & $89.81$ & $84.59$ & - & - & -  & - & - & - & -  \\
			\hline
			Ours (fix$4\times30$) & R$50$+FPN  & $2$D  & F & $4 \times 30$  &   $95.27$ & $94.24$ & $86.83$  & $91.45$ & $89.95$ & $83.27$ & $85.24$ & $79.24$ & $64.92$  & $81.72$ & $74.75$ & $62.40$ & $57.42$  \\
			\hline
	 		Ours (fix$4\times20$) & R$50$+FPN   & $2$D  & F & $4 \times 20$    & $96.33$ & $94.32$ & $88.44$  & $92.56$ & $90.26$ & $84.29$ & $85.29$ & $79.47$ & $65.47$  & $82.12$ & $75.10$ & $63.08$ & $57.47$ \\
	 		\hline
 			Ours (fix$7\times7$) & R$50$+FPN   & $2$D  & F & $7 \times 7$    & $96.17$ & $94.30$ & $86.63$  & $92.03$ & $90.28$ & $83.47$ & $81.58$ & $77.77$ & $62.83$  & $78.52$ & $73.59$ & $60.49$ & $50.82$ \\
	 		\hline
	 		Ours & R$50$+FPN  & $2$D  & V & $4 \times 30$    & $96.35$ & $94.87$ & $88.90$  & $92.13$ & $91.25$ & $84.74$ & $85.64$ & $80.45$ & $65.84$  & $82.21$ & $77.14$ & $63.55$ & $58.72$ \\
	 					\hline
			\end{tabular}
		}
	\end{center}
\end{table*}

{\color{black}{\subsection{Ablation Experiments}}}

In this section, a series of ablation experiments are carried out to analyze the design of each model part in detail. {\color{blue}{In ablation experiments, we use all real training images listed in Section~\ref{dataset} except COCO-Text during the fine-tuning process. Only the first and second data augmentation manners are adopted.}}

\subsubsection{\bf Joint Training vs.\ Separate Training}

{\color{black}{To validate the superiority of multi-task joint training,
we build a two-stage system (denoted as ``Ours (sep)'') in which the detection and recognition models are trained separately.
For fair comparison, the detector in ``Ours (sep)'' is built by
removing the recognition part from the original model and trained only with the detection loss.
As for recognition, we employ our 2D-attention based text recognition network~\cite{li2019aaai}, but train it without extra training data.
We can see from Table~\ref{Rebt:multi-task} that the two-stage system performs worse than the proposed model on ICDAR2013, ICDAR2015 and Total-Text. The results are consistent with that in our conference version~\cite{li2017towards}, which prove again that the multi-task joint training would result in much better model parameters for feature extraction and lead to better end-to-end performance. }}

\begin{figure}[!h]
	\begin{center}
		\includegraphics[width=0.48\textwidth]{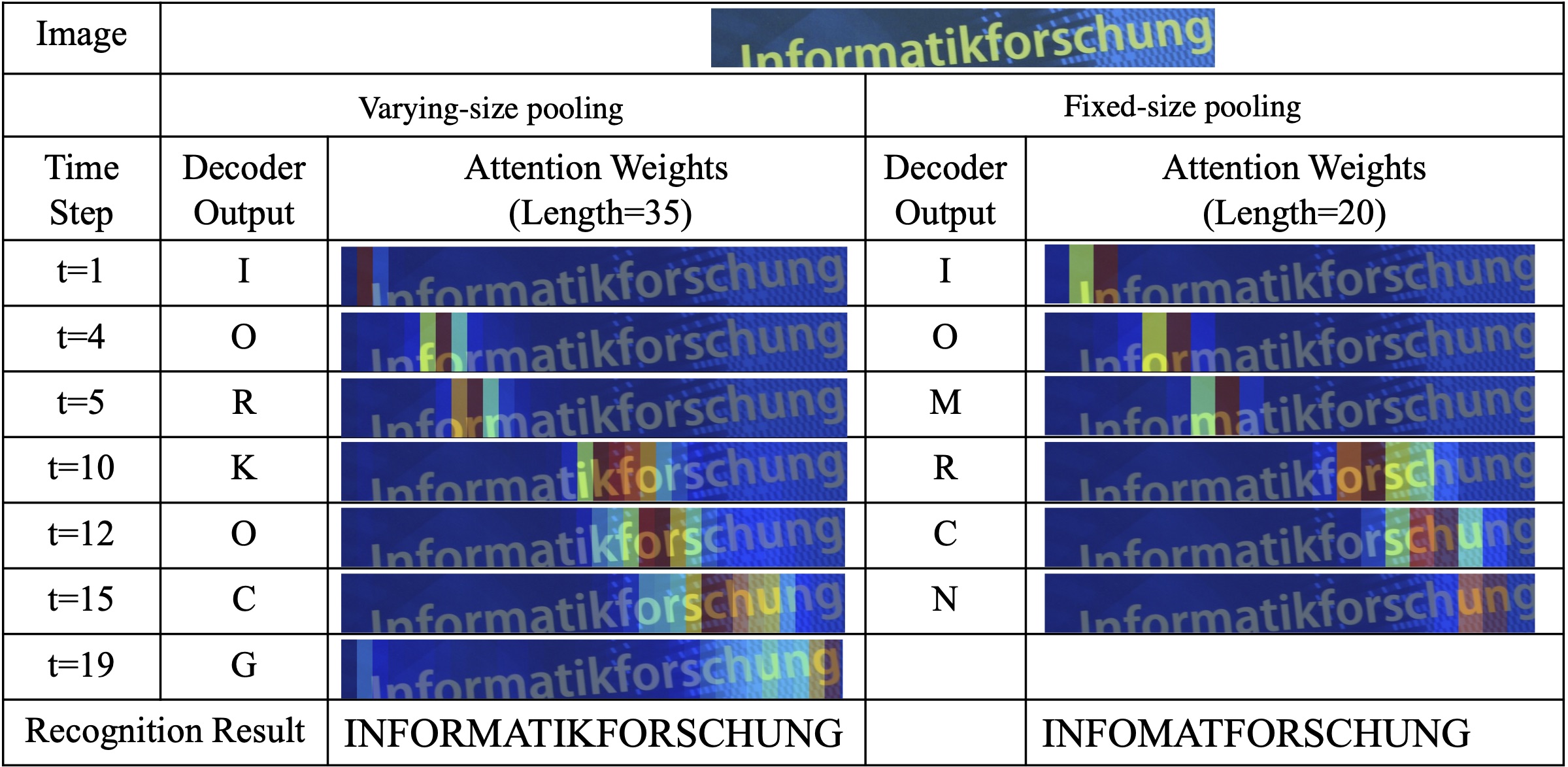}
	\end{center}
	\vspace{-0.3cm}
	\caption{Attention mechanism based sequence decoding process by varying-size and fixed-size RoI features separately. The heat maps show that at each time step, the position of the character to be decoded has higher attention weights, so that the corresponding local features are extracted and assist the text recognition. However, if we use the fixed-size RoI pooling, information may be lost during pooling, especially for a long word, which leads to an incorrect recognition result. In contrast, the varying-size RoI pooling preserves more information and leads to a correct result.}
	\label{fig:varyatt}
\end{figure}

\begin{table*}[!ht]
	\newcommand{\tabincell}[2]{\begin{tabular}{@{}#1@{}}#2\end{tabular}}
	\begin{center}
		\caption{\color{black}{Ablation study on RoI feature encoding
		methods.
		Compared to  using average pooling, LSTMs show superiority on sequential feature encoding.}}
		\label{Tab:fcpooling}
		\scalebox{0.88}{
			\begin{tabular}{l|c|c|c|c|c|c|c|c|c|c|c|c|c|c}
				\hline
				{\tabincell{c}{Model \\ Name}} & {\tabincell{c}{Encoding \\ in TDN }}    & \multicolumn{3}{|c}{\tabincell{c}{ICDAR2013 \\ Word-Spotting}} &  \multicolumn{3}{|c} {\tabincell{c}{ICDAR2013 \\ End-to-End}} &
				\multicolumn{3}{|c}{\tabincell{c}{ICDAR2015 \\ Word-Spotting}} &  \multicolumn{3}{|c|} {\tabincell{c}{ICDAR2015 \\ End-to-End}} & Total-Text
				 \\\cline{3-14} &&
				\multicolumn{1}{c|}{Strong} & \multicolumn{1}{|c|}{Weak}  & \multicolumn{1}{|c|}{Generic}   & \multicolumn{1}{c|}{Strong} & \multicolumn{1}{|c|}{Weak} & \multicolumn{1}{|c|}{Generic} 	&			\multicolumn{1}{c|}{Strong} & \multicolumn{1}{|c|}{Weak}  & \multicolumn{1}{|c|}{Generic}   & \multicolumn{1}{c|}{Strong} & \multicolumn{1}{|c|}{Weak} & \multicolumn{1}{|c|}{Generic} \\
			\hline
			Ours (AP) & AvePooling &   $94.13$ & $92.63$ & $85.06$  & $90.45$ & $89.49$ & $82.96$  & $83.62$ & $78.70$ & $62.85$  & $80.14$ & $74.99$ & $60.60$  & $55.95$  \\
			\hline
	 		Ours (AP+FC) & AvePooling+FC   & $94.23$ & $93.36$ & $87.61$  & $90.75$ & $89.76$ & $83.56$ & $83.62$ & $78.80$ & $63.34$  & $80.34$ & $74.82$ & $60.76$ & $57.06$ \\
			\hline
	 		Ours & LSTMs   & $96.35$ & $94.87$ & $88.90$  & $92.13$ & $91.25$ & $84.74$ & $85.64$ & $80.45$ & $65.84$  & $82.21$ & $77.14$ & $63.55$ & $58.72$ \\
			\hline
			\end{tabular}
		}
	\end{center}
\end{table*}

{\color{black}{Furthermore, we compare the detection performance of the two-stage model and the jointly trained model. Note that for ICDAR2013, both detection results are achieved without referring to recognition results, while for ICDAR2015 and Total-Text, attention weights obtained during recognition phrase are used to rectify the detected bounding boxes.
The detection results in Table~\ref{Tab:det} demonstrate that { the recognition loss in model training can also benefit the detection performance. } The proposed end-to-end model produces detection performance (F-measures) averagely $4\%$ higher than that given by ``Ours (sep)''. The detection results on ICDAR2013 are comparable with the state-of-the-art under three evaluation criteria. Nevertheless, they are worse than state-of-the-arts on irregular text datasets. Note that our work mainly focuses on the end-to-end text spotting scenario and only uses circumscribed rectangles as ground-truth bounding boxes for training. We leave the accurate text localization to future work.
}}

\subsubsection{\bf Fixed-size vs.\ Varying-size RoI Pooling}

Another contribution of this work is a varying-size RoI pooling mechanism, to accommodate the large variation of text aspect ratios. To validate its effectiveness, we compare the performance of models with varying-size RoI features ($H=4$ and $W_{max}=30$) and fixed-size ones. Different RoI pooling sizes are also tested for comparison.

{\color{black}{Experimental results in Table~\ref{Tab:roi} indicate that adopting varying-size RoI pooling improves the text spotting performance.  After collecting statistics of the aspect ratios of bounding boxes in the training data, we set the width of RoI to the medium value of $20$ in our comparison (the model is denoted as ``Ours (fix$4\times20$)''). It achieves the best text spotting performance among all fixed settings, but is still worse than using varying-size RoI pooling, with an $0.5 \%$ drop on F-measures. We visualize the attention heat maps based on varying-size RoI features and fixed-size RoI features respectively. As shown in Figure~\ref{fig:varyatt}, fixed-size RoI pooling may lead to a large portion of information loss for long words.

To peer off the impact of RoI pooling size adopted on model performance, we also test the model that uses $W=30$ in the fixed setting (denoted as ``Ours (fix$4\times30$)''). The F-measures drop about $1 \%$ comparing with the results by using ``Ours (fix$4\times20$)''. This result indicates that a longer $W$ adopted in the fixed setting may not be beneficial, as most image features will be distorted and stretched in that case. The deformation would be more serious for short words. These experiments also demonstrate the flexibility of our varying-size pooling method in dealing with the large diversity of text aspect ratios.

In addition, we test the model that pools RoI features to $7 \times 7 $ (named as ``Ours (fix $7\times7$)''), which is originally used in Faster-RCNN~\cite{renNIPS15fasterrcnn}. F-measures also decrease on all test datasets compared to the proposed model, especially on Total-Text.

}}

\begin{figure}[!b]
	\begin{center}
		\includegraphics[width=0.48\textwidth]{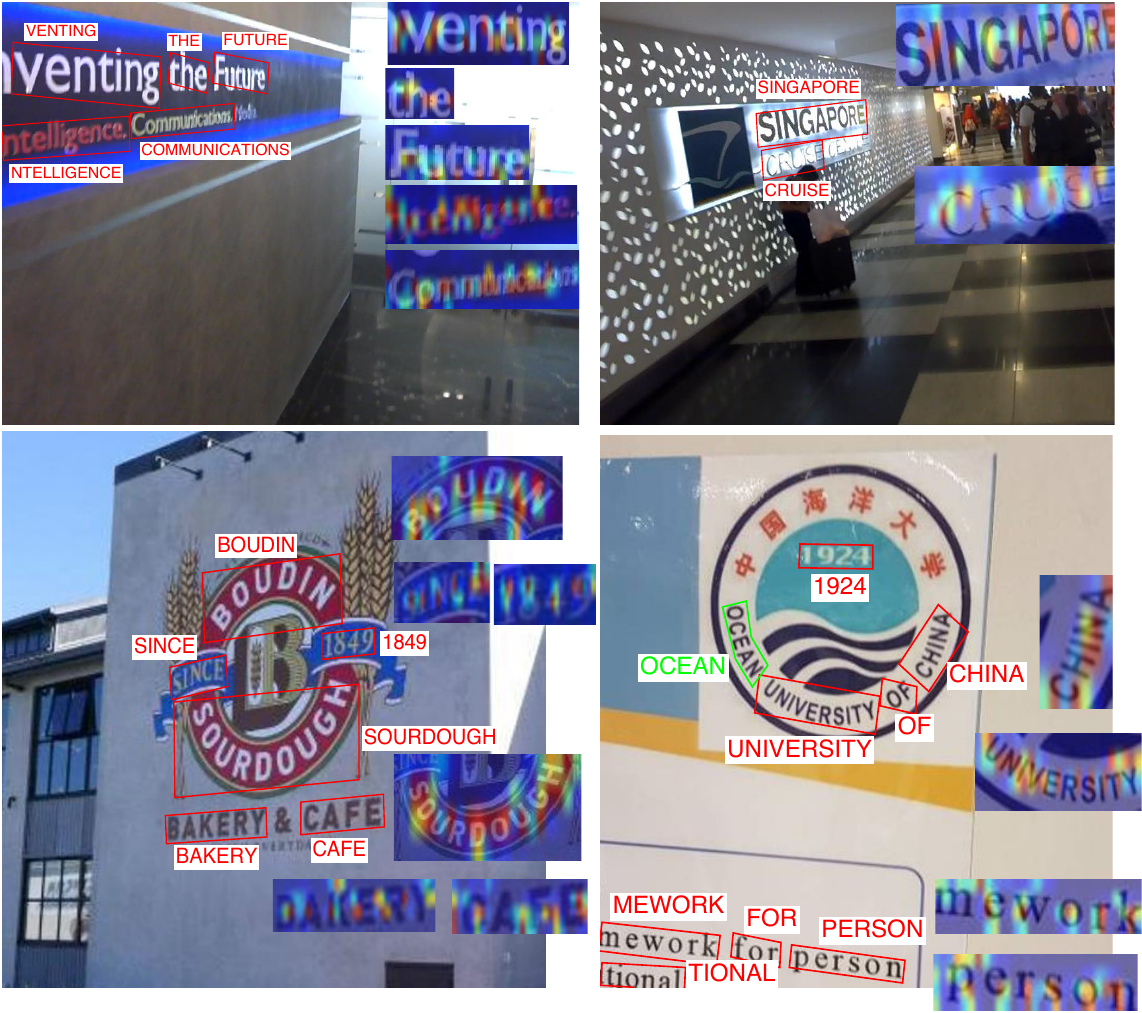}
	\end{center}
	\vspace{-3mm}
	\caption{Visualization of $2$D attention heat map for each word proposal by aggregating attention weights at all character decoding steps. The results show that the $2$D attention model can approximately localize characters, which provides assistance in both word recognition and bounding box rectification. Images are from ICDAR2015 in the first row and Total-Text in the second row. The red bounding boxes are both detected and recognized correctly. The green bounding boxes are missed words.
	}
	\label{fig:imattention}
\end{figure}

\begin{table*}[!ht]
	\newcommand{\tabincell}[2]{\begin{tabular}{@{}#1@{}}#2\end{tabular}}
	\begin{center}
		\caption{\color{black}{Ablation experiments on model architecture. ``Former (full)~\cite{li2017towards}'' shows the results from previous conference version. ``RNN Enc.'' shows whether TDN and TRN share $1$ layer of RNN encoder or not. ``Box Ref.'' means whether performing box refinement. Experiments are conducted not only on regular text dataset, but also on irregular ones for comprehensive evaluation. }}
		\label{Table:backbone}
		\scalebox{0.78}{
			\begin{tabular}{l|c|c|c|c|c|c|c|c|c|c|c|c|c|c|c|c|c}
				\hline
				{\tabincell{c}{Model \\ Name}} & {\tabincell{c}{Backbone }}  & Attn. & {\tabincell{c}{RNN \\ Enc.}} & {\tabincell{c}{Box \\ Ref.}} & \multicolumn{3}{|c}{\tabincell{c}{ICDAR2013 \\ Word-Spotting}} &  \multicolumn{3}{|c} {\tabincell{c}{ICDAR2013 \\ End-to-End}} &
				\multicolumn{3}{|c}{\tabincell{c}{ICDAR2015 \\ Word-Spotting}} &  \multicolumn{3}{|c|} {\tabincell{c}{ICDAR2015 \\ End-to-End}} & Total-Text
 				\\\cline{6-17} &&&&&
				\multicolumn{1}{c|}{Strong} & \multicolumn{1}{|c|}{Weak}  & \multicolumn{1}{|c|}{Generic}   & \multicolumn{1}{c|}{Strong} & \multicolumn{1}{|c|}{Weak} & \multicolumn{1}{|c|}{Generic} 	&			\multicolumn{1}{c|}{Strong} & \multicolumn{1}{|c|}{Weak}  & \multicolumn{1}{|c|}{Generic}   & \multicolumn{1}{c|}{Strong} & \multicolumn{1}{|c|}{Weak} & \multicolumn{1}{|c|}{Generic} \\
				\hline
			Former (full)~\cite{li2017towards} & VGG-16   & $1$D & Share & N  & $94.16$ & $92.42$ & $88.20$  & $91.08$ & $89.81$ & $84.59$ & - & - & -  & - & - & - & - \\

			\hline
	 		Ours ($1$D) & R$50$+FPN   & $1$D & Share & N  & $95.26$ & $93.94$ & $88.13$  & $91.93$ & $90.17$ & $84.26$  & $80.76$ & $76.03$ & $61.14$  & $77.40$ & $73.38$ & $59.08$ & $49.85$  \\
			\hline
			Ours ($2$D) & R$50$+FPN   & $2$D & Share  & N & $96.87$ &    $95.15$ & $88.06$  & $92.61$ & $91.05$ & $84.18$  & $82.12$ &  $77.70$ & $61.96$  & $78.75$ & $74.43$ & $60.49$ & $50.39$ \\
	 			\hline
			Ours (Shr) & R$50$+FPN  & $2$D & Share & Y & $96.87$ & $95.15$ & $88.06$  & $92.61$ & $91.05$ & $84.18$ &  $85.70$ & $80.14$ & $65.61$  & $82.43$ & $76.35$ & $63.30$ & $58.20$\\
 			\hline
		 	Ours & R$50$+FPN   & $2$D & Sep. & Y  & $96.35$ & $94.87$ & $88.90$  & $92.13$ & $91.25$ & $84.74$ & $85.64$ & $80.45$ & $65.84$  & $82.21$ & $77.14$ & $63.55$ & $58.72$ \\
			\hline
			\end{tabular}
		}
	\end{center}
\end{table*}

\subsubsection{\bf Effect of RoI Encoding Manner}

In the proposed framework, LSTMs are adopted to convert the varying-length RoI features into a fixed-size for the following text detection and recognition networks. Instead of using LSTMs, here, we extract a fixed-size holistic feature by average pooling across RoI features. A performance degradation of nearly $2 \%$ on F-measures is received comparing the results of ``Ours (AP)'' and ``Ours'' in Table~\ref{Tab:fcpooling}. Furthermore, we test the model with an extra $1024$D Fully-connected layer after average pooling (named as ``Ours (AP+FC)'') and find a slight improvement. However,  ``Ours'' still perform significantly
better than ``Ours (AP+FC)''.
These experiments illustrate the effectiveness of LSTMs on sequential feature encoding.

{\color{black}{\subsection{Improvements over %
Our Preliminary Results in
\cite{li2017towards}
}

To be specific, there are four major improvements on the model architecture over the previous conference version~\cite{li2017towards}, i.e., backbone network, attention structures, box refinement,
and the re-arrangement of RNN encoders. In this subsection, we will demonstrate the effect of each part in detail, so as to better understand our model.
}}

\subsubsection{\bf Effect of Backbone Network}

The conference version~\cite{li2017towards} used VGG-$16$ as the backbone network. Only the final convolutional layer is adopted for RoI feature extraction. In contrast, the new model adopts ResNet$50$ with FPN as the backbone, which extracts RoI features from different levels of feature pyramid according to their scales. As compared in Table~\ref{Table:backbone} between ``Former (full)~\cite{li2017towards}'' and ``Ours ($1$D)'', the new backbone framework gives F-measures gain around $1 \%$ on ICDAR2013, mostly because of a higher recall.

\subsubsection{\bf $1$D vs.\ $2$D Attention}
By comparing the results between ``Ours ($1$D)'' and ``Ours ($2$D)'', we find that using $2$D Attention instead of $1$D attention gives an roughly $1\%$ improvement to accuracy.
It is worth to note that even for the horizontal text in ICDAR2013, the $2$D attention mechanism is still better than the $1$D counterpart.
The reason may be caused by the more accurate character-level feature extraction during decoding process.
We also visualize the $2$D attention heat maps in Figure~\ref{fig:imattention}. Although trained in a weakly supervised manner (which means that it is trained without character-level annotations), the  attention model can approximately localize each character to be decoded, which, on one hand, extracts local feature for character recognition, on the other hand, indicates character alignment for bounding box refinement.

\subsubsection{\bf Box Refinement}
The proposed box refinement process is used together with
$2$D attention to boost text localization performance.
As shown in Table~\ref{Table:backbone}, our model with box refinement
(``Ours (shr)'') significantly outperforms that without box refinement (``Ours ($2$D)'') on the irregular text datasets
(roughly $3\%$ and $7\%$ improvement on ICDAR2013 and Total-Text respectively).
Box refinement is no longer needed for regular text, so the performance of ``Ours (shr)'' and ``Ours ($2$D)'' on ICDAR2013 is the same.
These results demonstrate that box refinement is a simple yet effective method for improving irregular text localization.

\begin{figure}[t!]
	\begin{center}
		\includegraphics[width=0.48\textwidth]{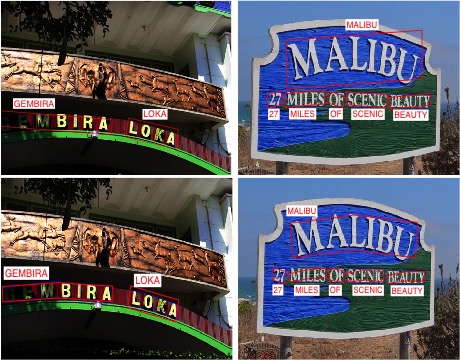}
	\end{center}
	\vspace{-2mm}
	\caption{Comparison between polygon and quadrangle fitting results on Total-Text.}
	\label{fig:Comtotal}
\end{figure}

{\color{blue}{Moreover, we have tried to fit 6-point polygons on Total-Text, according to the attention heat map. To be specific, we fit two quadrangles according to the attention weights in front and real half respectively. The two quadrangles are then connected and form a polygon for the word. The coordinates of the two junction points is the mean values of the closest coordinates from the two quadrangles.  Compared to the quadrangle results, the performance is indeed improved,
as demonstrated in Table~\ref{Tab:Abltotal}.  Visualization comparison is presented in Figure~\ref{fig:Comtotal}.
}}

\begin{table}[t!]
	\newcommand{\tabincell}[2]{\begin{tabular}{@{}#1@{}}#2\end{tabular}}
	\begin{center}
		\caption{Ablation experiments on box refinement manner. Fitting 6-point polygons on Total-Text can bring performance improvement of 0.4\% averagely on F-measures, compared to the quadrangle counterpart.}
		\label{Tab:Abltotal}
		\scalebox{0.95}{
			\begin{tabular}{l|c|c|c|c}
				\hline
				Method & \multicolumn{3}{|c}{\tabincell{c}{Detection}} &  \multicolumn{1}{|c} {\tabincell{c}{ End-to-End}}  \\\cline{2-5} 	& \multicolumn{1}{c|}{Recall} & \multicolumn{1}{|c|}{Precision}  & \multicolumn{1}{|c|}{F-measure}   &  \multicolumn{1}{c}{F-measure}  \\
				\hline
				Ours\_Quad & $60.23$ & $62.76$ & $61.47$  & ${58.72}$   \\
				\hline
				Ours\_Poly  & $60.75$ & $63.05$ & $61.88$  & ${59.11}$   \\
				\hline
			\end{tabular}
		}
	\end{center}
\end{table}

\subsubsection{\bf Sharing of LSTM encoders}
{\color{black}{In the conference version~\cite{li2017towards}, a layer of LSTM encoder with $1024$ hidden states is shared between TDN and TRN. Another layer of LSTMs with $1024$ units is adopted in TRN.
While in the new model, the LSTM encoders are separated in TDN and TRN. One layer of LSTM encoder with $1024$ states are applied in TDN, and two layers of LSTM encoder with $512$ states are used in TRN.
This modification leads to almost $4$M less parameters and
$20$ms speeding-up (on Titan X), without significantly affecting the accuracy (comparing ``Ours (shr)'' and  ``Ours'' in Table~\ref{Table:backbone}).

\begin{figure*}[ht!]
	\begin{center}
		\includegraphics[width=0.8\textwidth]{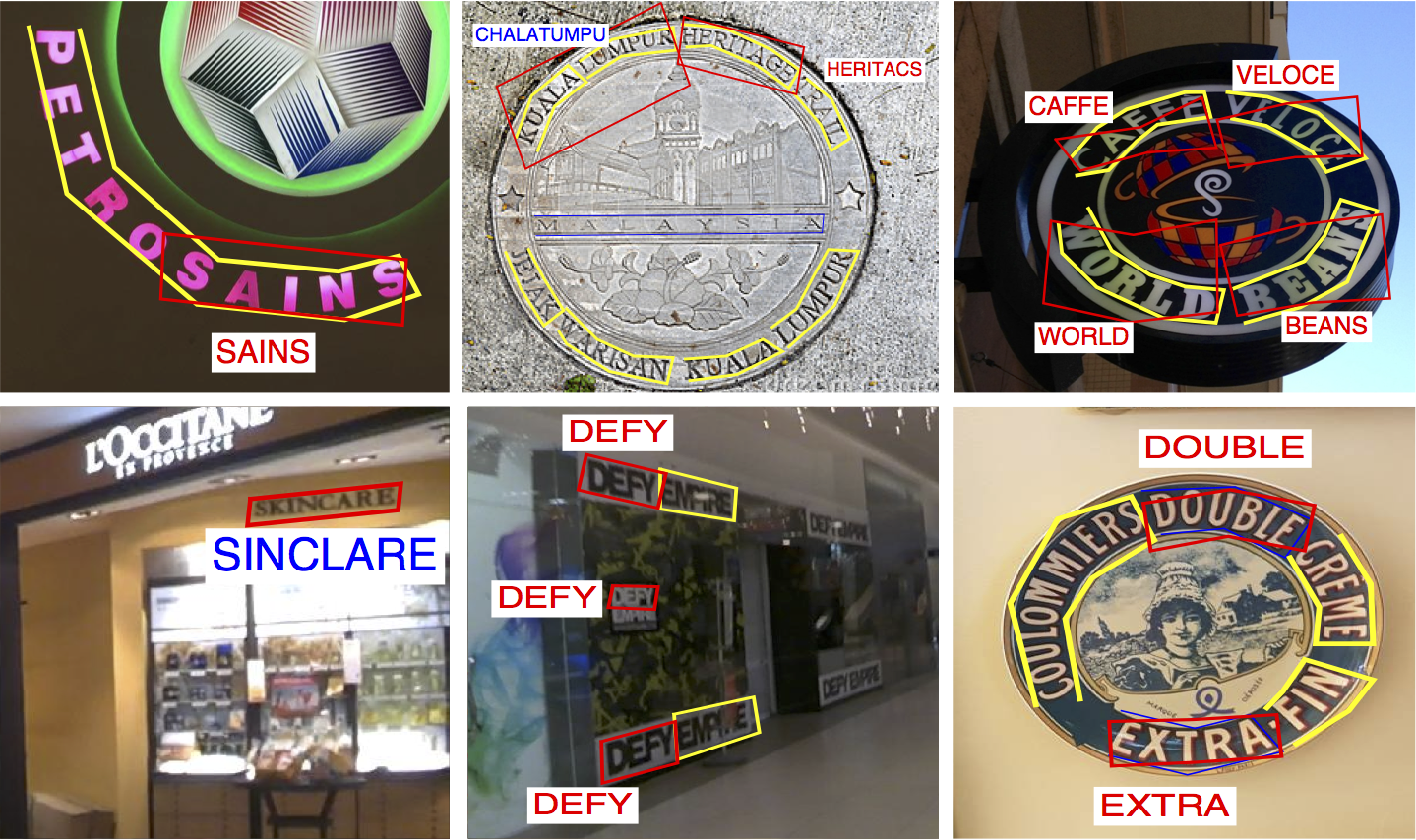}
	\end{center}
	\caption{Failure cases of our model. The red bounding boxes and labels are our detection and recognition results. The yellow ones are missed detection, which cannot be correctly recognized. The blue labels are wrongly recognized, although been well detected.
	}
	\label{fig:fail}
\end{figure*}

\subsubsection{\bf Speed}
Owing to hyper-parameter tuning (including reducing $W_{max}$ from $35$ to $30$, reducing the size of two FC layers in TDN from $2048$ to $1024$, and the above LSTM encoder adjustment) and a better implementation (replacing \texttt{for} loops with GPU-friendly code), the overall running speed of the new model
is around $2$ times faster than the conference version ($0.5$s \textit{vs.}\  $1$s for processing a $720 \times 1280$ image on a Titan-X GPU).}}

{\color{blue}{\subsection{Failure Cases Analysis}

The failure of text spotting can be caused either by %
inaccurate
detection results or by %
false recognition results.  If the text is not detected or the bounding box only covers part of text, %
recognition is doomed to fail.
In addition, it %
sometimes fails to recognize the word even it has been detected correctly.
As presented in Figure~\ref{fig:fail}, there are a variety of reasons for the failure, such as text that appears in a low contrast against the background,%
text that is of a very small size, uncommon fonts or blurred. Our work is also %
incapable of
spotting vertical text. Moreover, the heuristic box refinement process based on the attention  maps is not perfect. As indicated in Figure~\ref{fig:imattention}, the heatmap may not hit in the middle of each character, but drifts to the upper or lower side.
It may sometimes fail to rectify the bounding box.
This is much room for improvement.
}}

\section{Conclusions}

In this paper we have presented a %
simple
end-to-end trainable network for simultaneous text detection and recognition in natural scene images.
A novel RoI encoding method has been proposed, considering the large diversity of aspect ratios of word bounding boxes. We use a $2$D attention model that  is capable of indicating character locations accurately, which assists word recognition as well as text localization. Being robust to different forms of text layouts, our approach performs
well for both regular and irregular scene text.

For future work, one potential direction is to use convolutions or self-attention to %
replace %
the recurrent networks used in the framework, so as to speed up the computation.
Our current framework may fail to recognize text that is aligned vertically, which
deserves further study.

\section*{Acknowledgment}
P. Wang's participation in this work was in part supported by the National Natural Science Foundation of China (No.\ 61876152, No.\ U19B2037).

\ifCLASSOPTIONcaptionsoff
  \newpage
\fi

\bibliographystyle{IEEEtran}
\bibliography{IEEEabrv,mybibfile}

%
\end{document}